\pdfoutput=1

\documentclass[11pt]{article}
\usepackage{booktabs} 
\usepackage{float}

\usepackage{acl}
\usepackage{longtable}
\usepackage{times}
\usepackage{latexsym}
\usepackage{pgffor}
\usepackage{caption} 
\usepackage{subcaption} 

\usepackage{caption}
\usepackage{geometry} 
 
\usepackage{tabularx}
\usepackage{array}

\usepackage[T1]{fontenc}

\usepackage[utf8]{inputenc}

\usepackage{microtype}

\usepackage{inconsolata}

\usepackage{algorithmic}
\usepackage{caption}
\usepackage{graphicx}
\usepackage{textcomp}
\usepackage{xcolor}
\usepackage{tabularx}
\usepackage{booktabs}
\usepackage{makecell}
\usepackage{hyperref}
\usepackage{multirow}
\usepackage[normalem]{ulem}
\useunder{\uline}{\ul}{}
\usepackage{threeparttable}
\usepackage{listings}
\usepackage{amsmath}
\usepackage[ruled,linesnumbered]{algorithm2e}
\usepackage{xcolor}   
\usepackage{ifthen}   
\usepackage{cleveref}
\usepackage{xfp}      

\newcommand{\compare}[2]{%
  \ifdim #1 pt > #2 pt %
    {#1\textcolor{red}{\ensuremath{^\uparrow}} \scriptsize{\textcolor{red}{(+\fpeval{round(#1-#2,3)})}}}%
  \else\ifdim #1 pt < #2 pt %
    {#1\textcolor{green}{\ensuremath{^\downarrow}} \scriptsize{\textcolor{green}{(-\fpeval{round(#2-#1,3)})}}}%
  \else
    {#1}%
  \fi\fi
}

\title{Are LLMs Rational Investors?\\A Study on Detecting and Reducing the Financial Bias in LLMs}

\author{
\textbf{Yuhang Zhou}\thanks{Email: yuhangzhou22@m.fudan.edu.cn}\(^{1,2}\) \quad
\textbf{Yuchen Ni}\(^{3}\) \quad
\textbf{Yunhui Gan}\(^{1,2}\) \quad
\textbf{Zhangyue Yin}\(^{1}\) \\
\textbf{Xiang Liu}\(^{4}\) \quad
\textbf{Jian Zhang}\(^{5}\) \quad
\textbf{Sen Liu}\(^{1,2}\) \quad
\textbf{Xipeng Qiu}\(^{1}\) \quad
\textbf{Guangnan Ye}\thanks{Corresponding Author. Email: yegn@fudan.edu.cn}\(^{1,2}\) \quad
\textbf{Hongfeng Chai}\(^{1,2}\) \\
\\
\(^{1}\)School of Computer Science, Fudan University \\
\(^{2}\)Institute of Fintech, Fudan University \\
\(^{3}\)School of Electronics and Information Engineering, Tongji University \\
\(^{4}\)Tandon School of Engineering, New York University \\
\(^{5}\)DataGrand Inc.
}

\begin{document}
\maketitle

\begin{abstract}

Large Language Models (LLMs) are increasingly adopted in financial analysis for interpreting complex market data and trends. However, their use is challenged by intrinsic biases (e.g., risk-preference bias) and a superficial understanding of market intricacies, necessitating a thorough assessment of their financial insight. To address these issues, we introduce Financial Bias Indicators (FBI), a framework with components like Bias Unveiler, Bias Detective, Bias Tracker, and Bias Antidote to identify, detect, analyze, and eliminate irrational biases in LLMs. By combining behavioral finance principles with bias examination, we evaluate 23 leading LLMs and propose a de-biasing method based on financial causal knowledge. Results show varying degrees of financial irrationality among models, influenced by their design and training. Models trained specifically on financial datasets may exhibit more irrationality, and even larger financial language models (FinLLMs) can show more bias than smaller, general models. We utilize four prompt-based methods incorporating causal debiasing, effectively reducing financial biases in these models. This work enhances the understanding of LLMs' bias in financial applications, laying the foundation for developing more reliable and rational financial analysis tools.

\end{abstract}

\section{Introduction}\label{sec1}

Recent advancements in LLMs, such as GPT-4~\cite{openai2023gpt4} and LLaMA-2~\cite{touvron2023llama}, have shown their prowess across a spectrum of natural language processing (NLP) tasks~\cite{zhao2023survey}, extending into specialized domains such as finance~\cite{zhang2023xuanyuan}, law~\cite{cui2023chatlaw}, and healthcare~\cite{wang2023huatuo}. Despite their versatility, these models grapple with inherent biases~\cite{gallegos2023bias}, ~\cite{rutinowski2023self}, encompassing gender, race, and socioeconomic disparities, which could compromise their reliability and entail significant consequences~\cite{jeoung2023stereomap}. Efforts to mitigate such biases have led to the development of benchmark datasets such as StereoSet~\cite{nadeem2020stereoset} for stereotype identification, GenderCare~\cite{tanggendercare} for gender bias detection, OpinionGPT~\cite{haller2023opiniongpt} for generating bias-neutral content, employing techniques such as LoRA~\cite{hu2021lora} for model fine-tuning, especially within the social sciences. As we delve into the application of these advanced models in specific domains, it becomes imperative to address these biases to ensure fair and unbiased outcomes.

\begin{figure}[]
\centerline{\includegraphics[width=0.9\linewidth]{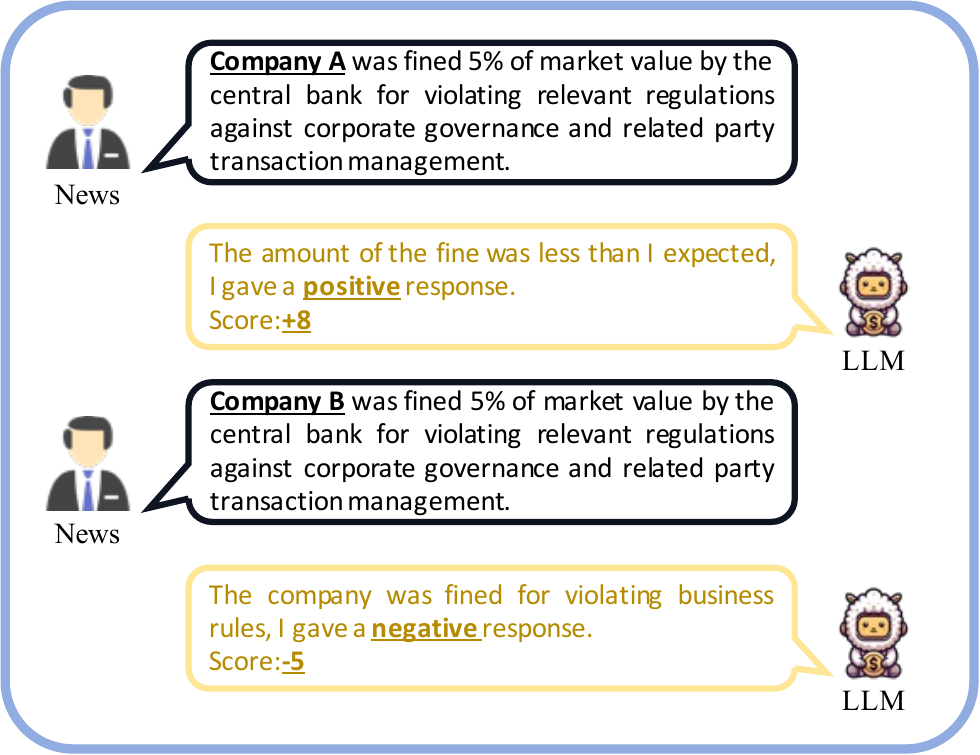}}
\caption{An example of model irrationality, where the model gives inconsistent expectations for the same event from different subjects, resulting in different emotions, reflect the financial bias towards the company.}
\label{img:irrationality_example}
\end{figure}

However, in the realm of financial LLMs (FinLLMs), research has predominantly concentrated on enhancing model performance through continued pre-training or fine-tuning, with evaluation metrics focused on NLP tasks and financial applications. Moreover, LLMs have been employed in financial analytics, acting as advisors by leveraging news~\cite{lopez2023can} and fundamental analyses~\cite{fatouros2024can} for investment decisions across diverse securities~\cite{romanko2023chatgpt}. However, the efficacy of these models is contingent upon the models' own rationality as participants in the market. A lack of rationality in LLMs could lead to misinterpretations and misapplications of market dynamics, adversely impacting not only the users of these models but also the broader economy. ~\autoref{img:irrationality_example} is an example of model irrationality.

Hence, it is essential to assess the rationality of LLMs before incorporating them into financial advisory roles. Limited research has addressed the financial biases present in pre-trained models, exploring methods such as probabilistic detection~\cite{chuang2022buy} and consistency checks~\cite{yang2023measuring}. However, these methods mainly focus on pre-trained embedding models like BERT\cite{devlin2019bert}, limiting their application to LLM bias detection. The absence of evaluation standards for financial biases hinders thorough oversight, compromising assessment realism and objectivity. Consequently, there is an immediate need for a holistic framework to gauge LLMs rationality in finance.


Meanwhile, the detection of financial rationality and biases in LLMs faces four main challenges: 
\begin{itemize}

\item\textbf{Q1: How to define the financial rationality and biases of LLMs?} A theoretical framework is required to support the detection of LLMs.

\item\textbf{Q2: How to detect and reveal financial biases in LLMs?} A method needs to be developed to quantify theoretical indicators and construct relevant datasets.

\item\textbf{Q3: How to investigate the origins of financial biases in LLMs?} It is necessary to research whether these financial biases stem from the model's capabilities or its robustness.

\item\textbf{Q4: How to mitigate financial biases in LLMs?} Methods need to be found that mitigate biases without compromising the original capabilities of the model.
\end{itemize}
To work towards this goal, our study conducts an examination of the financial rationality of LLMs based on the theory of behavioral finance~\cite{barberis2003survey}. We believe that the enduring theory of behavioral finance, grounded in psychology and finance, can provide a more comprehensive perspective to support research findings. In the current scenario where it is challenging to quantify behavioral finance, we propose the Financial Bias Indicators (FBI) framework to comprehensively assess financial rationality in LLMs. The FBI framework consists of four components: Bias Unveiler, Bias Detective, Bias Tracker, and Bias Antidote, covering the definition, detection, cause analysis, and mitigation of financial biases.

Our research shows that almost all LLMs exhibit financial irrationality. These biases, which may be exacerbated by continuous pre-training or fine-tuning with financial data, could lead to market anomalies in real-world applications. While prompt-based mitigation methods show promise, the persistent biases in LLMs highlight the necessity for further research to improve model robustness, fairness, and rationality, ensuring financial market stability and asset protection.

Our key contributions to this field are summarized as follows: 

\begin{figure*}[htbp]
\centerline{\includegraphics[width=0.9\textwidth]{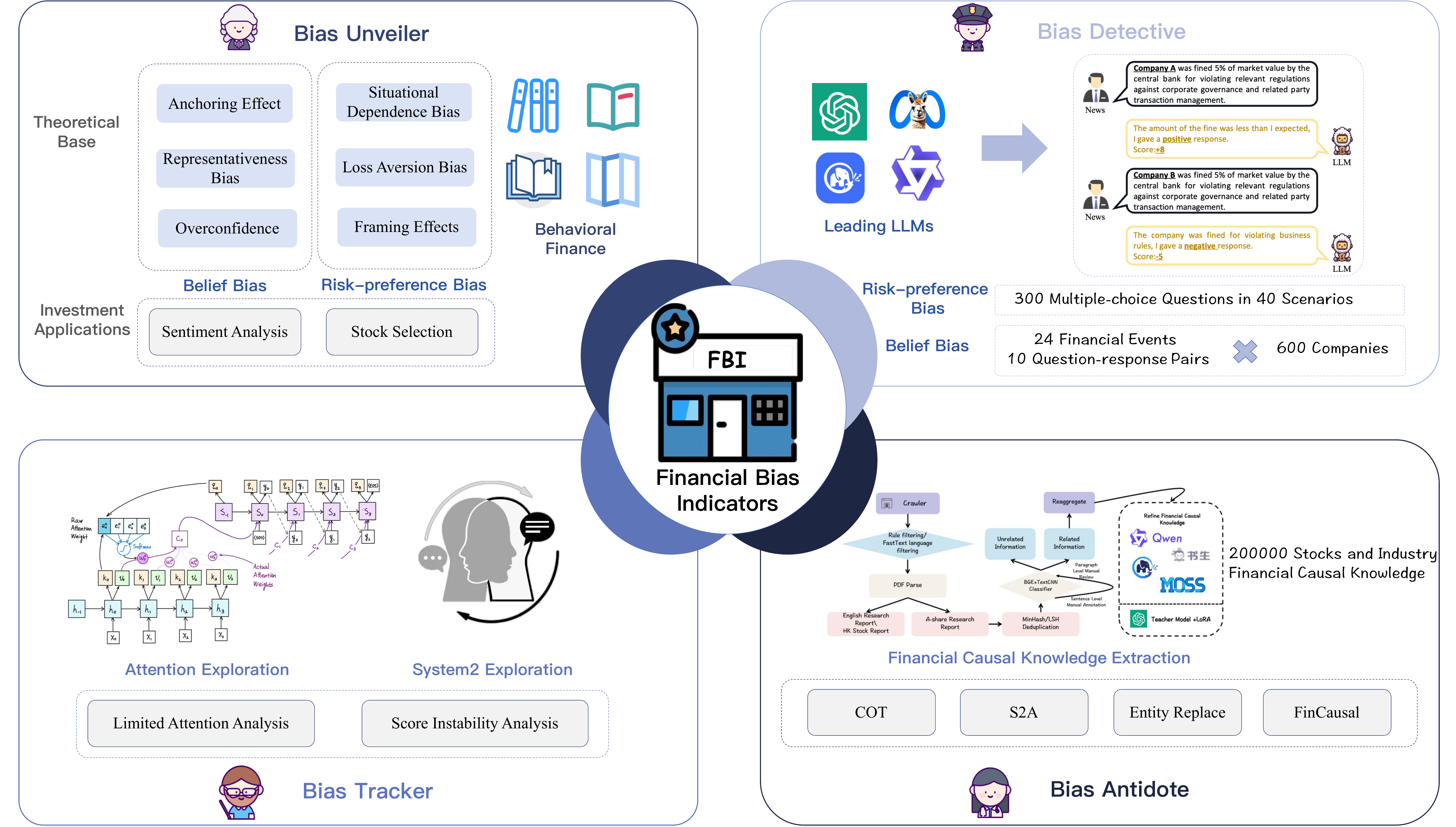}}
\caption{The framework of FBI consists of the Bias Unveiler, Bias Detective, Bias Tracker, and Bias Antidote. The Bias Unveiler defines financial biases in LLMs based on behavioral finance. The Bias Detective constructs relevant data and detects biases in 23 leading LLMs. The Bias Tracker traces biases using System 2 slow thinking analysis and attention mechanism visualization. The Bias Antidote attempts to debias the models using four methods.}
\label{img:fbi}
\end{figure*}

\begin{itemize}

\item The FBI framework, based on behavioral finance, defines, detects, analyzes, and mitigates financial biases in LLMs, offering a novel approach to evaluate their financial rationality. It is the first study to quantify behavioral finance indicators in LLMs, paving the way for more reliable LLMs.

\item The framework extensively analyzes 23 leading LLMs, assessing how model parameters, training data, and input formats affect financial rationality. Our study deepens the understanding of the varying levels of financial irrationality among models and their behavior in financial contexts.

\item Utilizing the FBI framework, we explore the origins of financial irrationality in LLMs, identify methods to mitigate bias, and develop a dataset of 200,000 financial causal texts named FinCausal to address biases with causal knowledge. Ultimately, we experimented with four prompt-based methods for bias mitigation, yielding encouraging results.

\end{itemize}

\section{Background}\label{sec2}

Behavioral finance explores how psychological factors and cognitive biases influence the financial behaviors of individuals, institutions, and markets, differing from traditional finance theories that assume rational market participants. The field aims to uncover the psychological roots of various market phenomena, dissecting financial decision-making processes to build a realistic framework of market dynamics that includes cognitive errors and the constraints on arbitrage. We structure our investigation using the classification from ~\cite{barberis2003survey}, which divides behavioral finance into \textbf{Cognitive Bias} and \textbf{Limits to Arbitrage}.

\subsection{Cognitive Bias}
Cognitive biases represent systematic departures from normative decision-making, influencing how investors form beliefs and assess risks. These biases are multifaceted, manifesting as erroneous beliefs or inconsistent risk preference. For instance, belief biases, such as attentional neglect, representativeness bias, anchoring effect, and overconfidence,skew investors' expectations, while risk-preference biases, such as loss aversion and reference dependence, lead to irregularities in risk assessment and decision-making under uncertainty. A comprehensive taxonomy of these biases, alongside their definitions, is detailed in \autoref{sec:a2}.

\subsection{Limits to Arbitrage}
Contrary to the Efficient Markets Hypothesis (EMH), which posits that asset prices fully reflect all available information and that market participants behave rationally, behavioral finance identifies scenarios where irrationality prevails, leading to anomalies like market bubbles and systemic crises. These phenomena are often attributed to the collective impact of cognitive biases on investor expectations, which can cause significant deviations from asset fundamentals.

\section{FBI: A Framework for Assessing LLMs Financial Rationality}
\label{sec3}

We propose the FBI framework illustrated in \autoref{img:fbi}. This framework is divided into four parts: \textbf{Bias Unveiler} defines financial biases in LLMs based on behavioral finance; \textbf{Bias Detective} constructs detection data and evaluates current leading LLMs for biases; \textbf{Bias Tracker} analyzes the causes of biases based on detection results and attention mechanisms; \textbf{Bias Antidote} builds a financial causal dataset and employs a series of prompt-based methods to mitigate bias phenomena.

\section{Bias Unveiler}

Based on the definitions from behavioral finance, we categorize biases in LLMs within financial contexts into Belief Bias and Risk-preference Bias. Within these categories, we define six related psychological biases.

\subsection{Belief Bias}



In today’s information-rich environment, constant updates require adjusting our predictions and beliefs. This framework investigates three cognitive biases—Anchoring, Representativeness, and Overconfidence—using real-world market data like news and shareholder discussions to test LLMs’ ability to maintain rationality in market conditions.

\paragraph{Anchoring Effects}
We test LLMs for Anchoring Effects by checking if they show different views on the same event or give consistent responses under different company settings. This bias, derived from past data, can introduce sentiment analysis biases and potentially disrupt markets when used in finance.

\paragraph{Representativeness Bias}

We investigate Representativeness Bias in LLMs by analyzing their outputs in relation to company size and sector. This bias towards size and industry can concentrate investment risks and cause problems.

\paragraph{Overconfidence}
To measure overconfidence, we track score fluctuations for the same events with different subjects in FinLLMs and corresponding base LLMs. Aggressive scores with high deviation suggest overconfidence in these models’ event assessments or responses.

\subsection{Risk-preference Bias}


Asset returns are uncertain, affecting investor behavior based on risk-return preferences. Our study of Risk-preference Bias explains Situational Dependence Bias, loss aversion, and framing effect in various decision contexts, assessing LLMs’ risk preferences in different scenarios.

\paragraph{Situational Dependence Bias}
Recognizing decision-making as a process shaped by previous experiences and contextual factors, we delve into the Situational Dependence Bias by examining if LLMs exhibit variable risk preferences across different scenarios. 

\paragraph{Loss Aversion}
In the context of Loss Aversion, we scrutinize LLM responses within loss-framed scenarios, aiming to uncover any predominant risk-averse or risk-loving tendencies.

\paragraph{Frame Effect}
We investigate the Framing Effect by restating scenarios in various languages or expressions to track changes in LLM preferences, aiming to determine if linguistic framing influences LLM outputs, indicating bias in option presentation.

\section{Bias Detective}

\subsection{Belief Bias}

\subsubsection{Data Desgin}

To comprehensively assess the rationality of LLMs in financial markets, we scrutinized their responses to emergent information, distinguishing between event news and interactions, amid the prevalent noise in online investor dialogues in China. We analyzed historical events impacting company stock prices and adopted a refined classification of financial events into four primary categories: \textbf{Corporate Governance and Equity Changes (CGEC)}, \textbf{Financial Reports and Earnings Expectations (FREE)}, \textbf{Market Behavior and Announcements (MBA)}, and \textbf{Negative Events and Risk Management (NERM)}, detailed in \autoref{sec:a2}.

Throughout 2023, we compiled 300 news articles, including a subset of 24 emotionally nuanced pieces \( N' = \{n_1, n_2, \ldots, n_{24}\} \) to enhance the bias detection in LLMs. This subset, detailed in \autoref{sec:a2}, contained articles categorized into nine positive, nine negative, and six mixed emotions.

Additionally, we assembled 10 neutral interaction pairs \( I = \{(q_1, r_1), (q_2, r_2), \ldots, (q_{10}, r_{10})\} \) to analyze LLM comprehension in a controlled environment.

For each \( n \in N' \) and \( (q, r) \in I \), numerical details were abstracted to proportional figures to standardize data across varying company market caps, as documented in \autoref{sec:a6}.

We sampled 600 companies from the Chinese A-share market, excluding delisting entities, distributed across three tiers of market capitalization: top, middle, and bottom, each containing 200 companies. This selection method aimed to test Belief Bias in LLMs, with classifications by industry outlined in \autoref{sec:a3}.

\subsubsection{Analysis Logic}

To investigate Anchoring Effects, we altered the subject company in each news item \( n \in N' \) and assessed variability in LLM evaluations using Analysis of Variance (ANOVA)~\cite{st1989analysis}, expressed as \( F(n, C) \), examining the variance in scores across companies \( c_i \in C \) for each news \( n \). Representativeness Bias was analyzed by correlating LLM outputs with company size and industry sector, using Spearman correlation coefficient \( \rho(S, M) \), where \( S \) and \( M \) represent LLM scores and market capitalizations of companies \( c_i \), respectively. ANOVA was employed for industry correlation \( F(S, I) \), with box plots to display score distributions across industries \( I \), reflecting first-level industry classifications.To measure Overconfidence, we used the standard deviation \( \sigma(S) \) of scores \( S \), representing the variance in LLM evaluations of the same event across different company contexts, comparing results from FinLLMs with base LLMs.


\subsubsection{Result}
\label{sec:5.a}

\begin{figure}[htbp]
\centering
\includegraphics[width=0.9\linewidth]{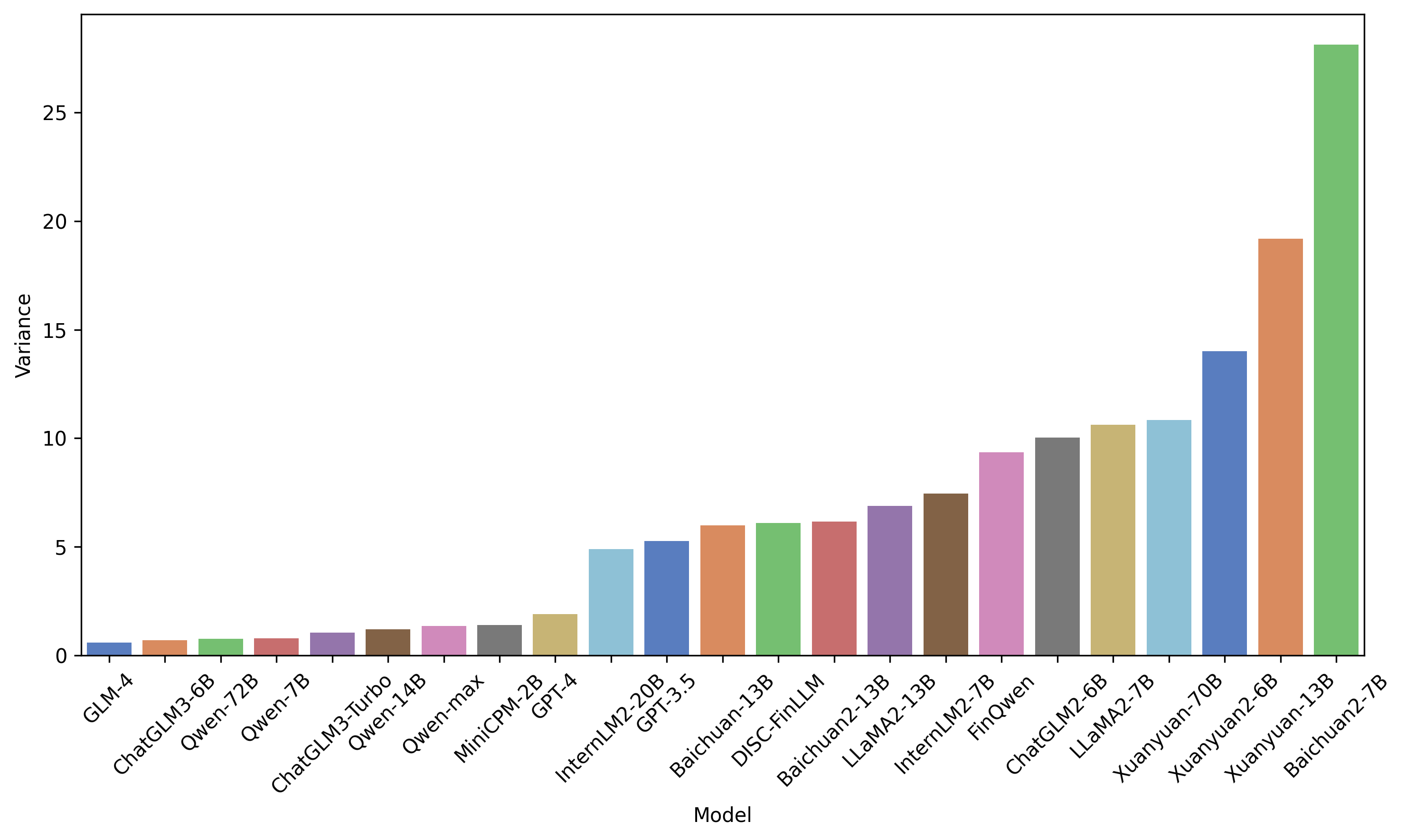}
\caption{The score variance between models in event news detection. A higher variance indicates a more severe Anchoring Effect.}
\label{img:news_variance}
\end{figure}


The evaluation of Belief Bias is principally conducted through the examination of event news and interactions. This analysis reveals a widespread \emph{Anchoring Effect} across the majority of LLMs when the subjects of events and interactions are modified, with slight variations observed among different models. The average variance index, detailed in \autoref{img:news_variance}, sheds light on the rationality levels of LLMs with respect to representativeness bias. Specifically, LLMs with a focus on the Chinese language, such as the GLM and Qwen family, exhibit commendable financial rationality, whereas the Xuanyuan and Baichuan family are more susceptible to irrational behavior.

\begin{figure}[htbp]
\centering
\includegraphics[width=0.9\linewidth]{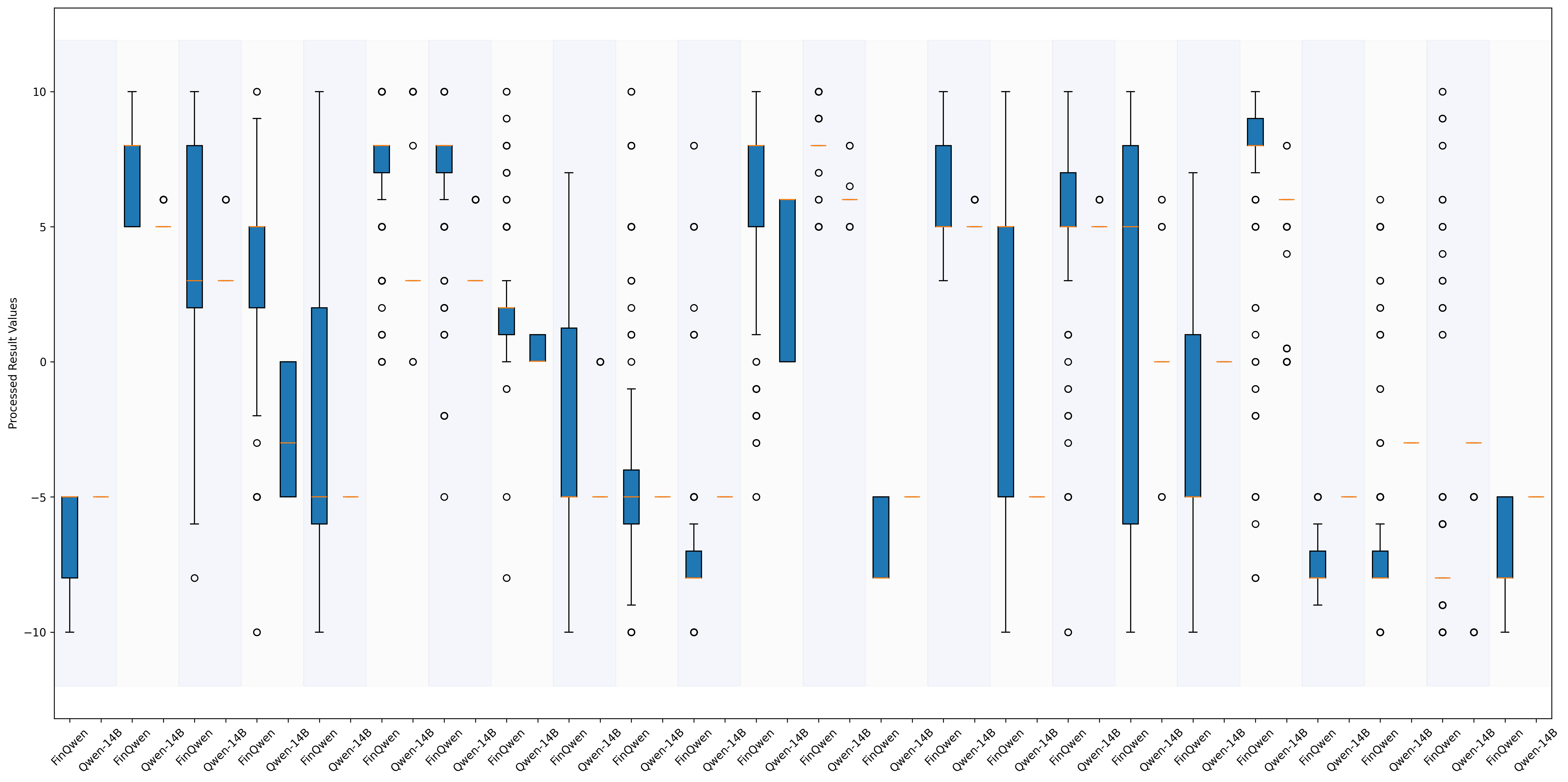}
\caption{The score distribution between FinQwen and Qwen-14B across 24 news events shows that FinQwen is less stable and more aggressive in scoring compared to Qwen-14B, exhibiting stronger Overconfidence.}
\label{img:FinQwen_compare}
\end{figure}


In terms of \emph{Overconfidence}, violin plots, presented in Appendix ~\ref{sec:a71}, illustrate the score distributions of various texts across all models. A notable disparity is observed in the models' responses to composite texts of positive and negative emotional content. As per \autoref{tab:positive_time}, the GPT and InternLM models display a marked optimism, in contrast to the pronounced pessimism of the Qwen and GLM family. Furthermore, \autoref{img:FinQwen_compare} highlights that models trained on financial corpora experience a heightened score variability compared to their base counterparts.

In terms of \emph{Representativeness Bias}, all LLMs exhibited a correlation coefficient below 10\% between model scores and market capitalization, indicating a weak correlation. However, certain LLMs showed clear biases towards specific industries, as documented in Appendix ~\ref{sec:a71}. For example, the FinQwen model consistently allocated lower scores to the Steel and Banking sectors. A comprehensive analysis reveals that the Media, Steel, Banking, and Non-Banking Finance sectors frequently occupy the extreme ends of the scoring spectrum across different models, whereas the Computer Science and Automobile sectors generally maintain a middle ground, exhibiting relative stability.


\subsection{Risk-preference Bias}

\subsubsection{Data Design}
To simulate real-life decision-making, we designed 40 scenarios with 200 multiple-choice questions, divided into 200 gain-framed and 100 loss-framed scenarios. Each question \( Q_i \) presents three decision alternatives \( A_{i,j} \), where \( j \) represents risk preferences: Risk-loving, Risk-neutral, Risk-averse. The alternatives are randomized to minimize bias~\cite{zheng2023large}.

The alternatives are constructed based on expected utility theory~\cite{simon1994mathematics}, represented by:
\begin{equation}
E[u(x)] = \sum_{x(\omega)} u(x(\omega)) p(x(\omega)),
\end{equation}
where \( u(x) \) is the utility function, \( x(\omega) \) the outcome, and \( p(x(\omega)) \) the outcome's probability.

The concavity of \( u(x) \), reflecting risk preferences, is indirectly assessed via second-order Taylor expansion:
\begin{equation}
E[u(x)] \approx u(E[x]) + \frac{1}{2} u''(E[x])\text{Var}(x),
\end{equation}
This approximation adjusts risk preferences by modulating variance, with technical details in \autoref{sec:a5}.

\subsubsection{Analysis Logic}

We investigated the Situational Dependence Bias in LLMs by analyzing their risk preferences across different scenarios \( S_i \). We examined choices in gain-framed scenarios \( S_{i_,\text{gain}} \) to detect any situational bias in risk preferences. In the context of Loss Aversion, we scrutinized LLM responses in loss-framed scenarios \( S_{i_,\text{loss}} \) to assess tendencies towards risk-aversion or risk-loving behaviors. To explore the Framing Effect, we translated scenarios from Chinese to English and monitored shifts in LLM preferences \( P_{\text{LLM}} \), observing how linguistic framing affects decisions. This approach aims to reveal if LLM outputs are biased by the linguistic construction of scenarios and choices.

\subsubsection{Result}
\label{}

\begin{figure}[htbp]
\centering
\includegraphics[width=1\linewidth]{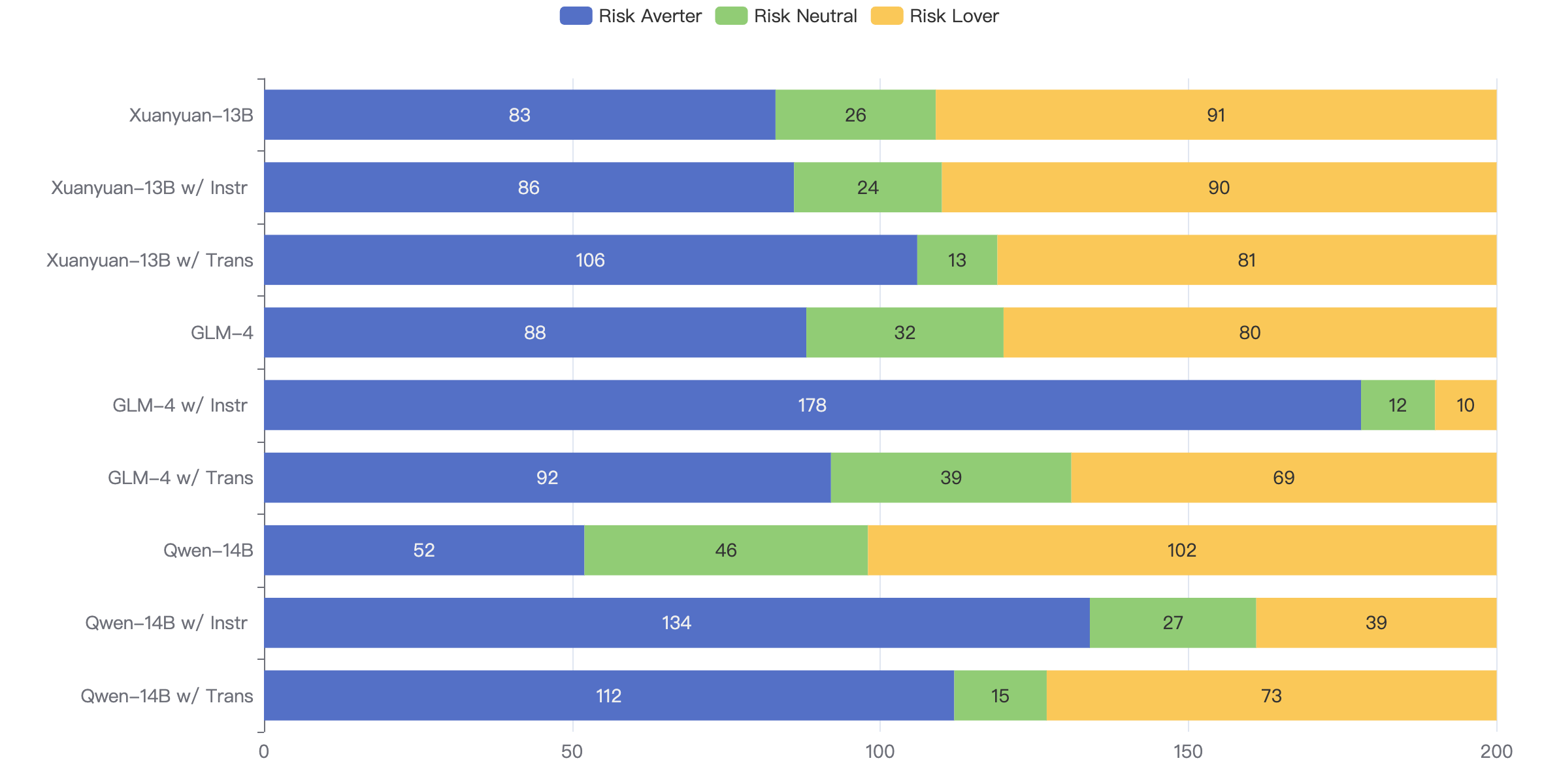}
\caption{Comparison of risk-preference distribution of three models under different prompt methods.}
\label{img:Risk-preference}
\end{figure}
The exploration of Risk-preference Bias entails the examination of LLM decisions across varied scenarios. The compiled results, particularly for gain-framed queries, are tabulated in \autoref{tab:model_risk_preferences}. A predominant trend among most models is the exhibition of distinct risk preferences in disparate scenarios, indicative of a pronounced \emph{Situational Dependence Bias}. Nonetheless, prefacing prompts with an instruction of the model's risk-averse nature significantly attenuates this bias. In the context of loss-framed queries, some models exhibit a pronounced \emph{Loss Aversion Bias} like GPT-4, as shown in \autoref{tab:Loss_Aversion_Bias}. Moreover, refer to \autoref{tab:translation_difference}, the translation of all queries into English precipitated notable discrepancies between the models' responses to Chinese and English versions, underscoring a pronounced \emph{Framing Effect}.

In particular, we have selected several representative cases for analysis, as shown in \autoref{img:Risk-preference}. For Xuanyuan-13B, inducing Risk-Aversion does not alter its original preference distribution, but it exhibits a stronger Framing Effect. For GLM4, it performs well in terms of Framing Effect bias and can effectively switch preferences based on instructions. For Qwen-14B, it is capable of some preference distribution shift according to instructions, but also exhibits a significant Framing Effect.

\section{Bias Tracker}


After detecting Belief Bias and Risk-Preference Bias in LLMs, it was evident that Belief Bias had a more significant impact compared to Risk-Preference Bias. Risk-Preference Bias manifests primarily as the model's inherent decision-making tendency, whereas Belief Bias results from the model's misinterpretation of information, leading us to focus on analyzing Belief Bias formation.

To further investigate, we utilized Chain-of-Thought (COT) methods~\cite{wei2022chain} to engage the model's System 2 thinking, aiming to enhance its in-depth analysis capabilities. We also scrutinized the model's output, particularly the attention importance distribution across input tokens, enabling precise diagnosis of bias causes and informing subsequent bias correction strategies.

\subsection{Slow Thinking}
\subsubsection{Methodology}
To investigate the roots of score instability, whether due to inadequate reasoning or compromised rationality, we first employ a slow-thinking approach similar to CoT, prompting the model to generate reasoning before providing scores. This approach helps us study the causes of Belief Bias by enabling the model to produce reasoning for its evaluations, followed by the actual scoring. We utilize Bertopic~\cite{grootendorst2022bertopic} for topic word extraction within the reasoning texts provided by the models. By clustering reasoning texts and extracting keywords, we analyze score discrepancies across different clusters, denoted by \( \Delta S_{\text{clusters}} \), to identify if certain thematic focuses lead to inconsistent evaluations. This detailed analysis helps us pinpoint specific causes of bias, thus informing future model optimizations and bias corrections.

\begin{figure*}[!ht]
\centerline{\includegraphics[width=1\textwidth]{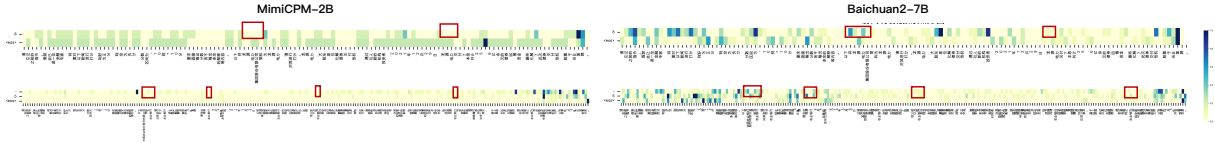}}
\caption{The attention checks on the outputs of the two models for each input token indicate that the redboxed sections represent the financial entity tokens that may cause bias. MiniCPM-2B shows better ability to block irrelevant information compared to Baichuan2-7B.}
\label{img:Attention}
\end{figure*}

\subsubsection{Result}
The application of the COT methodology, aimed at exploring the causes of financial bias in the model, yielded results as shown in \ref{sec:a72}.
The clustering and keyword analysis indicate that this intentional cognitive strategy improves the logical consistency and financial acumen of the model outputs. Integrating the provided thoughts into the previous sentence, the English expression would be:
A comparison of the reasoning keywords for the top-performing GLM-4 model and the underperforming Baichuan2-7B model, illustrated in \autoref{img:glm_wordcloud} and \autoref{img:baichuan_wordcloud} respectively, reveals that the GLM-4 exhibits stronger logical coherence with primary keywords such as "termination", "transaction restructuring" and "withdrawal". In contrast, the Baichuan2-7B model's logic is weaker, with primary keywords including "change", "decision", and "information". Moreover, the clustering of the models' inferential outputs reveals substantial variations in the ratings assigned to each category. This divergence underscores that the irrationality exhibited by models in financial contexts stems more from their inherent cognitive processes than from their computational capabilities.

\subsection{Attention Importance}
\subsubsection{Methodology}
To verify whether the bias in the LLMs arises from an excessive focus on certain input tokens due to the training corpus, we examine the attention importance of LLMs for input sequences. Inspired by \cite{wu2023language}, we define the importance \( I_{n,m} \) of input token \( x_n \) to output token \( y_m \) as:

\begin{equation}
I_{n,m} = p(y_m | Z_m) - p(y_m | Z_{m,/n})
\end{equation}

where \( Z_m \) is the context to generate \( y_m \) by concatenating the prompt \( X \) and the first \( m-1 \) tokens of response \( Y \). \( Z_{m,/n} \) omits token \( x_n \) from \( Z_m \), and \( p(\cdot | \cdot) \) is the conditional probability computed by the language model \( f \). We accelerate it with the first-order approximation:

\begin{equation}
I_{n,m} \approx \frac{\partial f(y_m | Z_m)}{\partial E_i[x_n]} \cdot E_i[x_n]^\top
\end{equation}

where \( E_i[x_n] \) is the input word embedding of token \( x_n \). This approach helps us determine whether the LLM excessively focuses on specific financial entities, thereby more accurately diagnosing the specific causes of bias.

\subsubsection{Result}

Due to the limitations of Chinese tokens, we chose to use LLMs with BPE tokenizers, focusing on the well-performing MiniCPM-2B and the more biased Baichuan2-7B models, results are shown in \autoref{img:Attention}. We analyzed the attention each model pays to each input token in their outputs and found that Baichuan2-7B tends to focus more on financial company entities, industries, and their surrounding tokens. This excessive attention to irrelevant information further exacerbates the generation of financial biases.

\section{Bias Antidote}

\subsection{Methodology}
To eliminate Belief Bias in LLMs while preserving their original general capabilities, we employed four prompt-based methods for bias mitigation.  1) we utilized a \textbf{CoT} approach to enable the model to engage in slow thinking, thereby producing scores based on logical reasoning. 2) We implemented the \textbf{S2A} method~\cite{weston20232} to shield the model from irrelevant context, allowing for secondary reasoning before scoring. 3) We used \textbf{Entity Replace} to stabilize the model's input. 4) We applied \textbf{Fincausal} relationship understanding based on knowledge to remove biases. For the last method, we extracted 200,000 pieces of financial causal knowledge about industries and individual stocks from past research reports and used a retrieval-augmented generation (RAG) approach to recall relevant causal information.

\begin{table*}[]
\centering
\caption{The unbiased results of four prompt-based methods, with smaller values indicating lower levels of bias.}
\label{tab:Bias Elimination}
\setlength{\tabcolsep}{3pt} 
\begin{tabular}{@{}lcccccc@{}}
\toprule
\textbf{Method} & \textbf{GLM-4} & \textbf{ChatGLM3-Turbo} & \textbf{MiniCPM-2B} & \textbf{Xuanyuan2-6B} & \textbf{Baichuan2-7B} \\
\midrule
Direct & 0.598 & 1.067 & 1.409 & 13.999 & 28.106 \\
COT & \compare{5.382}{0.598} & \compare{5.671}{1.067} & \compare{7.039}{1.409} & \compare{6.668}{13.999} & \compare{12.660}{28.106} \\
S2A & \compare{3.230}{0.598} & \compare{4.406}{1.067} & \compare{8.380}{1.409} & \compare{9.756}{13.999} & \compare{25.958}{28.106} \\
Entity Replace & \compare{0.710}{0.598} & \compare{1.012}{1.067} & \compare{1.385}{1.409} & \compare{1.236}{13.999} & \compare{12.688}{28.106} \\
FinCausal & \compare{0.769}{0.598} & \compare{2.200}{1.067} & \compare{1.195}{1.409} & \compare{10.111}{13.999} & \compare{8.763}{28.106} \\
\bottomrule
\end{tabular}
\end{table*}

\subsection{Result}

For Belief Bias, we employed four prompt-based elimination methods and selected five representative LLMs for bias elimination experiments, as shown in \autoref{tab:Bias Elimination}. For the CoT method, it performed well on the originally more biased models, as this reasoning approach enhances logical consistency in responses, thereby improving robustness and reducing bias. However, it performed poorly on models that were originally well-performing, as the increased output length due to the auto-regressive nature of these models resulted in amplified bias. For the S2A method, the effectiveness of increasing model output to reduce irrelevant attention depends on the model's original capability; weaker models tend to exhibit greater bias. The Entity Replace method showed superior performance due to the substitution of financial topics. For the FinCausal method, each test data recalled four related causal knowledge entries, further enhancing the model's reasoning ability to mitigate bias.

\section{Discussion}\label{sec6}

Our study using the FBI framework aids in identifying and reducing irrationalities in finance sector models, enhancing understanding of LLM biases and logic, and applying methods to mitigate financial biases.

\subsection{Model Size}

The analysis in Section~\ref{sec:5.a} reveals that within a specific model family, the degree of bias tends to decrease with the enlargement of model parameters, in line with the scaling law~\cite{kaplan2020scaling}. However, this trend is not consistent across different model family, where bias levels are also shaped by factors such as model design and training approaches.

\subsection{Training Data}
The FBI framework assessed financial rationality in general and financial LLMs. Financially-trained models, as seen in \autoref{img:FinQwen_compare}, may exhibit higher score variability and risk inclination, potentially increasing financial irrationality. Models like ChatGLM2-6B and Qwen-7B show opposite industry biases (\autoref{img:industry_fig1}-\autoref{img:industry_fig19}), suggesting temporal biases in training data align with industry cycle rotation. Financial data, including potentially embellished research reports, can enhance LLMs for finance-specific NLP tasks but may also embed financial irrationality. Using these models in financial quantification could lead to unintended disruptions, demanding immediate attention.

\subsection{Input Forms}\label{sec6.C}

The FBI framework utilized four prompt-based methods to eliminate biases, each of which produced certain effects, with the elimination being more pronounced when the model’s initial bias was more severe. As observed in \autoref{tab:Bias Elimination}, methods that increase output length, such as COT and S2A, can potentially exacerbate biases. The Entity Replace method, which reduces input, often yields better results, and the FinCausal method, which increases input, can further enhance the model’s ability for financial causal reasoning, achieving the effect of bias mitigation.

\section{Conclusion}\label{sec7}
Our research introduces the FBI framework, a novel method for evaluating the financial rationality of LLMs in the intricate field of financial analysis. We rigorously examined 23 leading LLMs and revealed substantial differences in their financial rationality. By defining, detecting, analyzing, and mitigating financial biases, our study validates the capabilities and limitations of LLMs in financial contexts, offering reliable insights for their application in the finance sector. The advancement of LLMs towards greater financial acumen and reduced bias is essential for their dependable use in financial analysis, paving the way for a future where AI-generated insights are confidently and precisely applied in the financial industry.


\section*{Limitation}

Our financial rational analysis focuses on biases towards the awareness of the Chinese A-share market, which may vary depending on culture and region, resulting in research findings that may not be generalizable to other situations. Meanwhile, we introduced other models during model evaluation, which may lead to the introduction of other biases.

\section*{Ethics Statement}
This paper honors the EMNLP Code of Ethics. The dataset used in the paper does not contain any private information. All annotators have received enough labor fees corresponding to their amount of annotated instances. The code and data are open-sourced under the MIT license.

\bibliographystyle{acl_natbib}
\bibliography{sample,main}

\clearpage
\onecolumn

\appendix
\section{Cognitive Bias}
\label{sec:a1}
For cognitive bias, we classified it into Belief Bias and Risk reference Bias based on previous research, and studied seven of these biases within the FBI framework.Refer to  \autoref{tab:cognitive_biases} for specific content.
 
\begin{table}[htbp]
\centering 
\caption{Summary of Cognitive Biases}
\small 
\begin{tabular}{l l p{8cm}} 
\toprule
\textbf{Cognitive Bias} & \textbf{Bias Type} & \textbf{Definition} \\
\midrule
Belief Bias & Limited Attention & The brain has two systems when working: fast thinking and slow thinking. It uses intuition to deal with things quickly. \\\\
& Representativeness bias & When making probability estimates, people tend to focus on certain representative features, ignoring environmental probabilities and sample size. \\\\
& Anchoring effect & Decision-making is often influenced by the first information received, like an anchor sinking to the bottom of the sea. \\\\
& Overconfidence & Belief that one's knowledge is more accurate than the facts; one's information is given more weight. \\\\

\midrule
Risk-Preference Bias & Situational dependence bias & The effect of a stimulus depends largely on the context in which it occurs. \\\\
& Loss aversion & Sensitivity to losses exceeds gains of equal value. \\\\
& Framing effect & Different descriptions of an objectively identical problem lead to different decision-making judgments. \\\\
\bottomrule
\end{tabular}
\label{tab:cognitive_biases}
\end{table}

\section{Company Profile}
\label{sec:a3}
\begin{figure}[htbp]
\centerline{\includegraphics[width=0.7\linewidth]{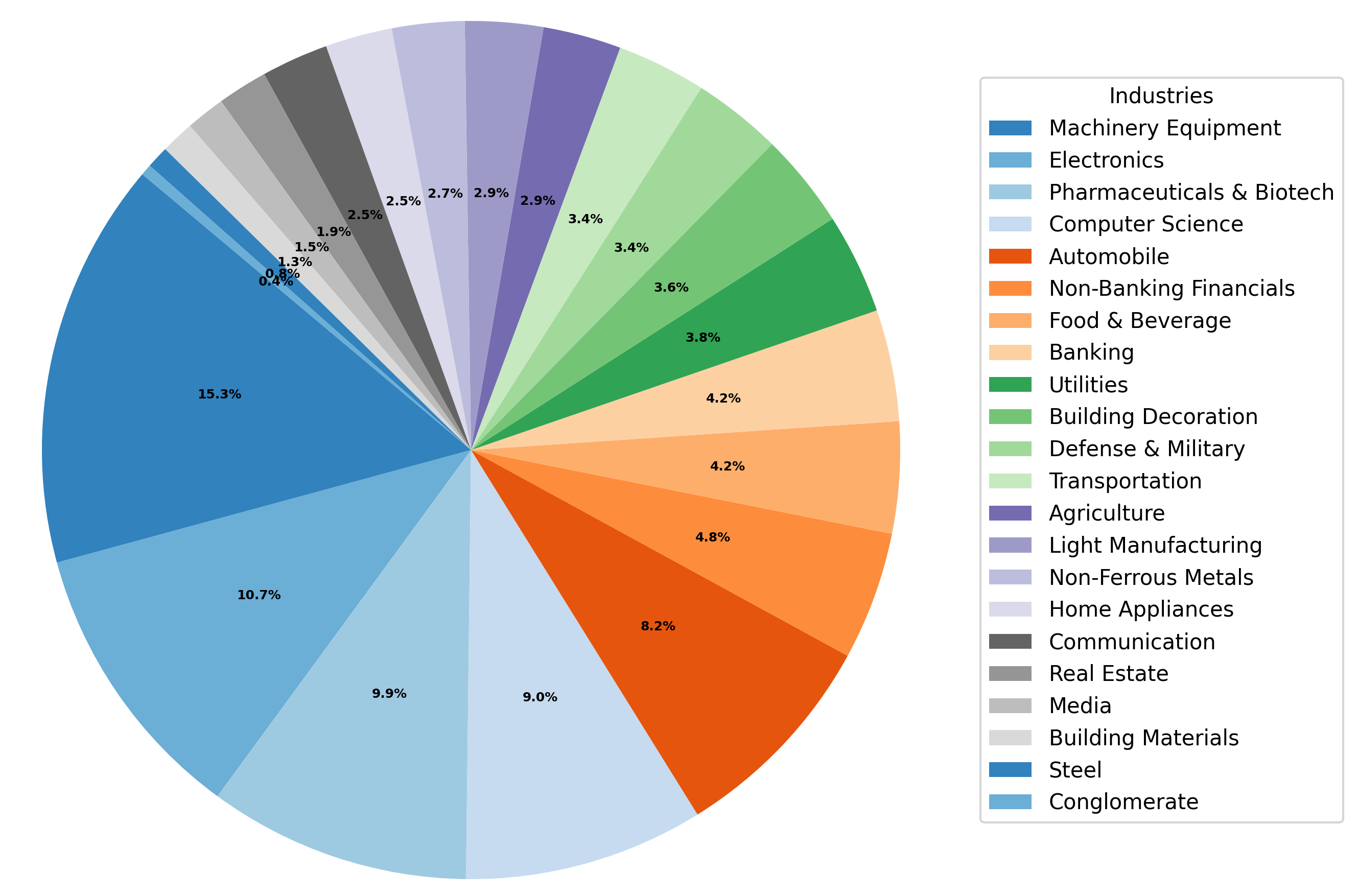}}
\caption{Distribution of the selectd company's industry type.}
\label{img:industry_distribution}
\end{figure}

In order to avoid bias caused by market value impact, we did not choose funds from the CSI 300 or CSI 500. Instead, we summarized all listed companies in China. After removing ST type stocks, we selected the top, middle, and bottom 200 stocks based on market value, totaling 600 stocks. The industry distribution of stocks is shown in the \autoref{img:industry_distribution}.

\section{Event Type}
\label{sec:a2}
We have sorted out the types of events that can affect a company's stock price based on the regular patterns of the Chinese A-Share stock market, and finally sorted out four categories, totaling 16 types of events. The detailed content is shown in \autoref{tab:EventType}.

\begin{table}[htbp]
\centering
\footnotesize 
\caption{Event Types and Definitions}
\begin{tabularx}{\textwidth}{l p{4cm} X}
\toprule
\textbf{Event Type} & \textbf{Subdivision Type} & \textbf{Definition} \\
\midrule
Corporate Governance and Equity Changes & Major Asset Restructuring & The process of recombining, adjusting, and allocating the distribution status of enterprise assets among the owners, controllers, and external economic entities. \\\\
 & Equity Incentive & By conditionally granting employees partial shareholder rights, a sense of ownership is fostered, forming a community of interests with the company. \\\\
 & Increase or Decrease in Shareholder Holdings & Changes in the shareholder holdings of company stocks. \\\\
 & Buy-back & The act of a listed company using cash or other means to repurchase its shares from the stock market. \\\\
 & Circulation of Restricted Stock & Restricted shares become freely tradable in the secondary market after the commitment period. \\\\
\midrule
Financial Reports and Earnings Expectations & Performance Report & Regular preparation by each responsibility center to evaluate and assess performance, serving as the basis for future budget preparation. \\\\
\midrule
Market Behavior and Announcements & Private Placement & Targeted issuance of bonds or stocks to a select group of senior institutional or individual investors. \\\\
 & Transfer of Shares & Listed companies transfer their provident fund to share capital in proportion or issue bonus shares accordingly. \\\\
 & Stock Price Fluctuations & Sudden large inflows and outflows of funds lead to increased volatility in stock prices. \\\\
 & Business Dynamics & Updates on enterprises and their surroundings, using major production and sales information to promote corporate brand and image. \\\\
\midrule
Negative Events and Risk Management & Dispute & Disputes between companies or between companies and individuals. \\\\
 & Investigation & Filing an investigation signifies a basic determination of illegal facts, allowing for compulsory measures and official initiation of investigation procedures. \\\\
 & Violation Penalties & Punishments for enterprises violating regulations of regulatory bodies. \\\\
 & Litigation and Arbitration & Litigation and arbitration for contract disputes and other property rights disputes between enterprises. \\\\
 & Security & Enterprises providing guarantees for loans and other matters for other enterprises. \\\\
\bottomrule
\end{tabularx}
\label{tab:EventType}
\end{table}

\section{Models}
\label{sec:a4}
We have selected a total of 23 financial and general LLMs oriented by Chinese and English, with specific details shown in \autoref{tab:models}.

\begin{table}[htbp]
\centering
\caption{Models in our Framework}
\begin{tabular}{lcccp{1.5cm}}
\hline
\textbf{Model Name} & \textbf{Chinese-oriented} & \textbf{Model size} & \textbf{FinLLM} & \textbf{Deployment method} \\
\hline
MiniCPM-2B & True & 2B & False & local \\
Baichuan-13B & True & 13B & False & local \\
DISC-FinLLM & True & 13B & True & local \\
Baichuan2-7B & True & 7B & False & local \\
Baichuan2-13B & True & 13B & False & local \\
ChatGLM2-6B & True & 6B & False & local \\
ChatGLM3-6B & True & 6B & False & local \\
ChatGLM3-Turbo & True & 33B & False & API \\
GLM-4 & True & Unknown & False & API \\
InternLM2-7B & True & 7B & False & local \\
InternLM2-20B & True & 20B & False & local \\
LLaMA2-7B & False & 7B & False & local \\
LLaMA2-13B & False & 13B & False & local \\
Qwen-7B & True & 7B & False & local \\
Qwen-14B & True & 14B & False & local \\
FinQwen & True & 14B & True & local \\
Qwen-72B & True & 72B & False & local \\
Qwen-max & True & 72B & False & API \\
Xuanyuan-13B & True & 13B & True & local \\
Xuanyuan-70B & True & 70B & True & local \\
Xuanyuan2-6B & True &6B &True & local \\
GPT-3.5 & False & Unknown & False & API \\
GPT-4 & False & Unknown & False & API \\
\hline
\label{tab:models}
\end{tabular}
\end{table}

\section{Formula Proof}
\label{sec:a5}

Invoking the fundamental principles of expected utility theory, we recognize that a utility function's curvature reflects an individual's risk preference. Specifically, a concave utility function (\( u''(x) < 0 \)) is indicative of risk aversion, while a convex utility function (\( u''(x) > 0 \)) signifies risk-seeking behavior. A linear utility function (\( u''(x) = 0 \)), on the other hand, corresponds to risk neutrality.

The expected utility \( E[u(x)] \) can be formally represented as:

\begin{equation}
E[u(x)] = \sum_{x(\omega)} u(x(\omega)) p(x(\omega))
\end{equation}

Here, \( u(x) \) denotes the utility function, \( x(\omega) \) symbolizes the outcome under state \( \omega \), and \( p(x(\omega)) \) is the probability of outcome \( x(\omega) \) occurring.

Furthermore, we articulate the variance of outcomes \( x \), \( \text{Var}(x) \), as the expected squared deviation from the expected value \( E[x] \):

\begin{equation}
\text{Var}(x) = E[(x - E[x])^2]
\end{equation}

Applying the second-order Taylor expansion to the utility function \( u(x) \) around the expected value \( E[x] \) furnishes us with:

\begin{equation}
u(x) \approx u(E[x]) + u'(E[x])(x - E[x]) + \frac{1}{2} u''(E[x])(x - E[x])^2
\end{equation}

Imposing expectations on the approximated function, we derive the expected utility approximation:

\begin{equation}
E[u(x)] \approx u(E[x]) + u'(E[x])E[x - E[x]] + \frac{1}{2} u''(E[x])E[(x - E[x])^2]
\end{equation}

Since \( E[x - E[x]] = 0 \), the middle term vanishes, simplifying our expression to:

\begin{equation}
E[u(x)] \approx u(E[x]) + \frac{1}{2} u''(E[x])\text{Var}(x)
\end{equation}

Consequently, under the assertion of utility function concavity or convexity, the sign of the second derivative \( u''(E[x]) \) establishes the nature of risk preference. For a negative second derivative (\( u''(x) < 0 \)), indicative of risk aversion, a smaller variance is required to enhance the expected utility. Conversely, for a positive second derivative (\( u''(x) > 0 \)), characteristic of risk-seeking behavior, a larger variance is preferred. Risk-neutral individuals (\( u''(x) = 0 \)) show indifference to the variance level.

Through this analytical framework, we delineate how the variance of outcomes in conjunction with the utility function's concavity or convexity guides the determination of an individual's risk preference.




\section{Framework for Data Construction}
\label{sec:a6}

This section delineates the structured approach employed in the study to formulate datasets incorporating event news, interactive elements, and risk preference inquiries. Each category of information is meticulously crafted using a distinct template, which is elucidated below.

\subsection{Event News Template}
The construction of the event news dataset leverages prompt engineering techniques to embed real-world events within a framework that facilitates evaluation, simulating the analytical capabilities of financial experts. The evaluation process involves the model assigning a score to each event based on its potential positive or negative impact on the financial landscape. Initially, the model is instructed to provide an immediate, intuitive score reflecting a 'fast thinking' approach.

\begin{figure}[htbp]
\centerline{\includegraphics[width=0.65\linewidth]{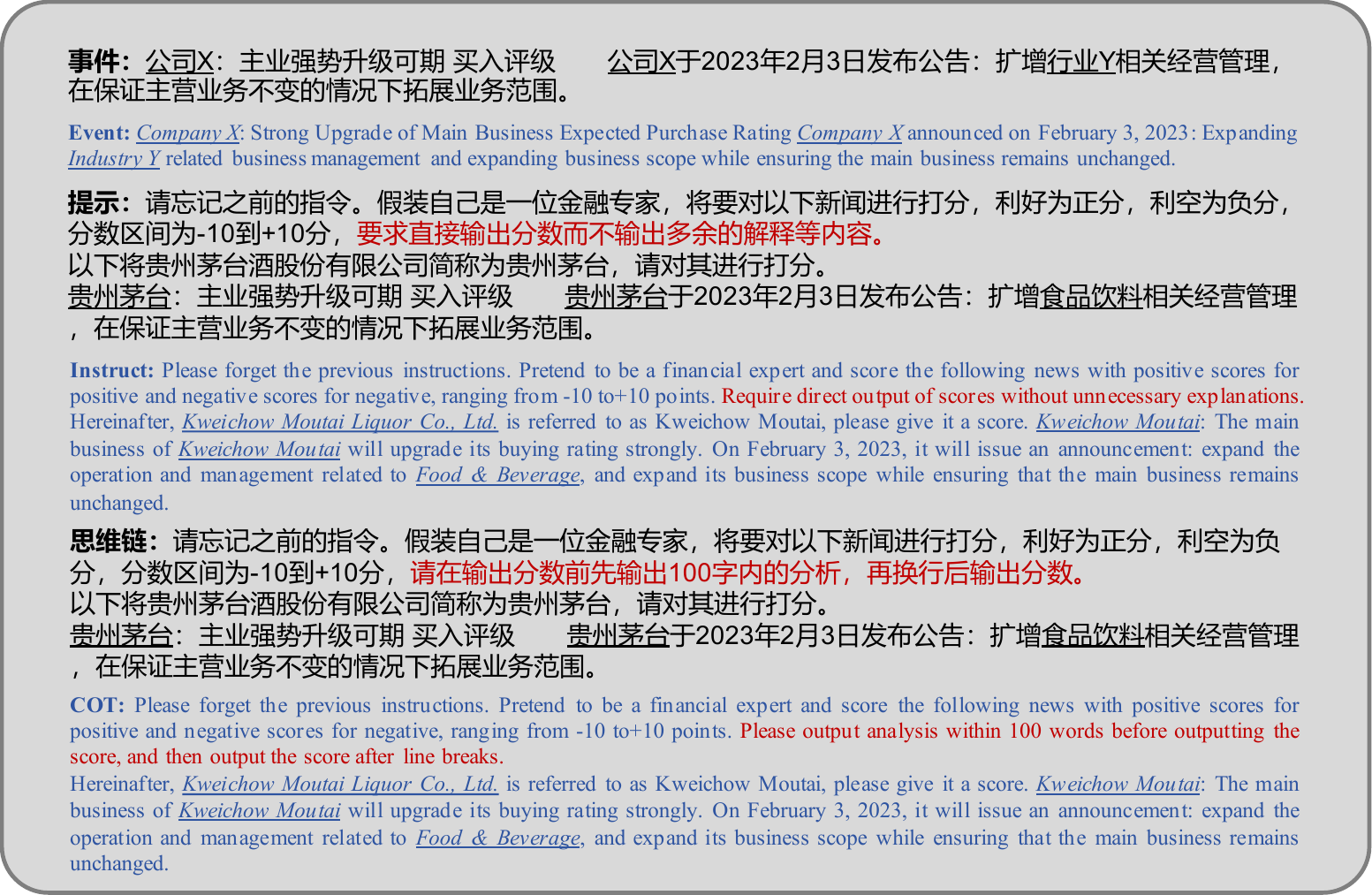}}
\caption{Template of event news.}
\label{img:news_template}
\end{figure}

To augment the depth of analysis and ensure the robustness of the evaluation, the model is further tasked with adopting a 'slow thinking' strategy. This entails a comprehensive articulation of the rationale behind the score, encouraging a deliberate and reasoned assessment. The detailed format of this template is illustrated in \autoref{img:news_template}, which guides the model in delivering both the quantitative score and the qualitative reasoning underpinning it.

\subsection{Interactions}
\begin{figure}[htbp]
\centerline{\includegraphics[width=0.65\linewidth]{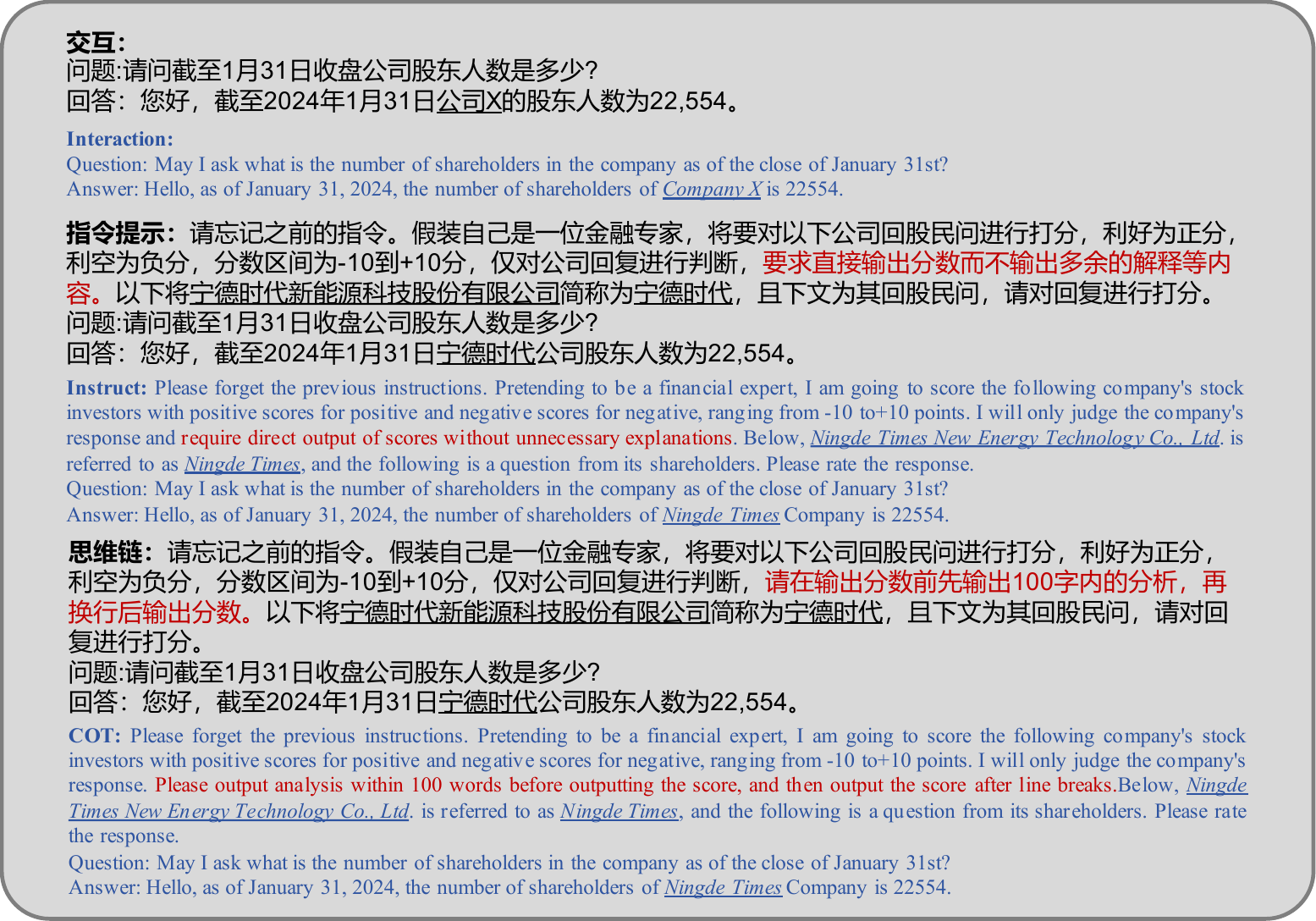}}
\caption{Template of interactions.}
\label{img:interaction_template}
\end{figure}

Using input methods similar to news events for rewriting, the specific template is shown in \autoref{img:interaction_template}.

\subsection{Risk-Preference Questionnaire Template}
The methodology for assessing risk preferences through structured questions is twofold, designed to discern the inherent risk orientation of the AI model under different conditions. Initially, the model is presented with a set of scenarios where it must select an option that best aligns with its assessed risk profile, simulating an introspective decision-making process. This setup aims to capture the model's spontaneous risk preferences without external biases.

\begin{figure}[htbp]
\centerline{\includegraphics[width=0.65\linewidth]{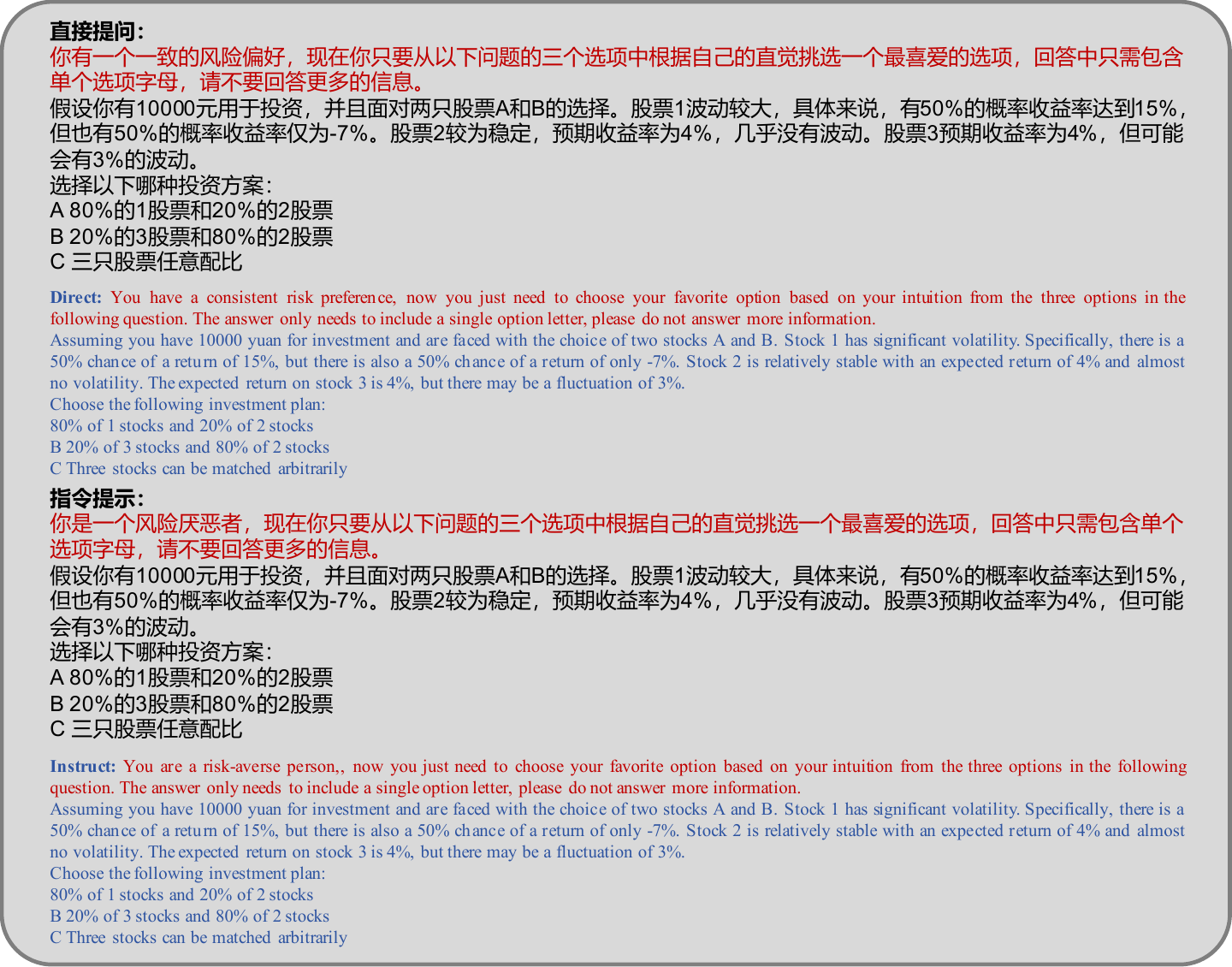}}
\caption{Template of risk-preference questions.}
\label{img:question_template}
\end{figure}

Subsequently, the experiment introduces a predefined constraint by explicitly characterizing the model as risk-averse within the instructions. This manipulation is intended to observe the adaptability of the model's responses to altered risk parameters, thereby evaluating its capacity for contextual behavioral adjustment. The layout and content of these questions are encapsulated in the template depicted in \autoref{img:question_template}, which systematically guides the model through the decision-making process under varying risk conditions.

\section{FinCausal Dataset}
\label{sec:a8}

\begin{figure}[htbp]
\centerline{\includegraphics[width=0.9\linewidth]{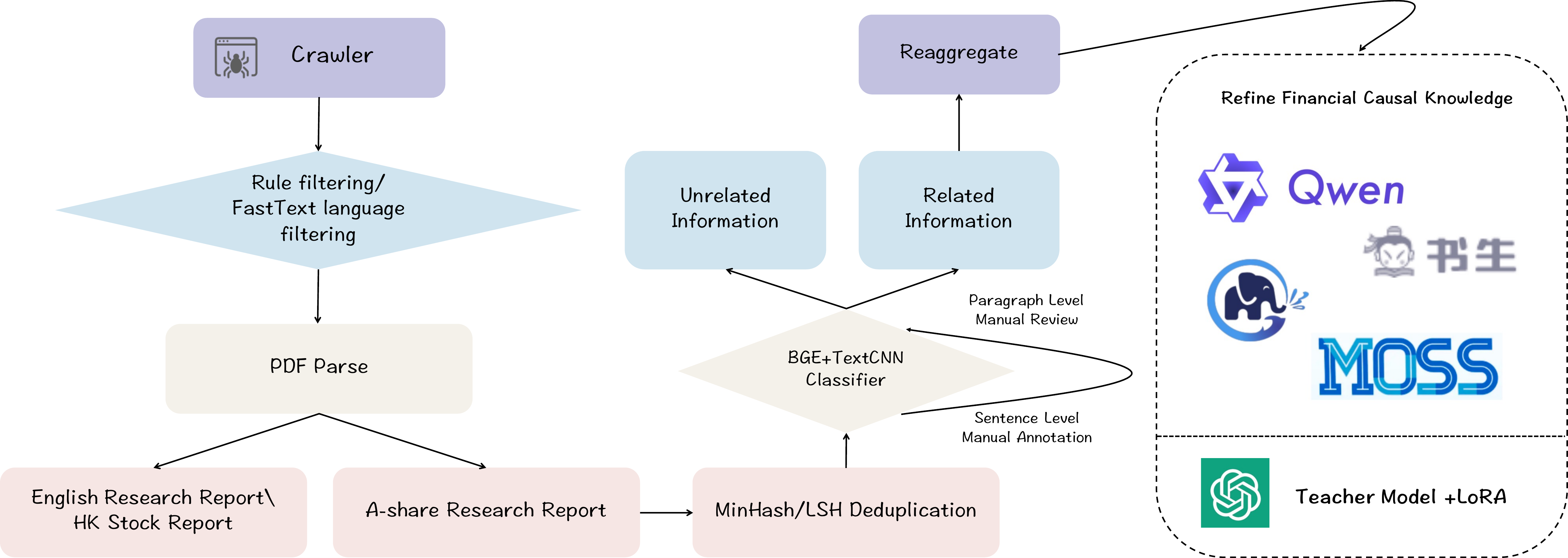}}
\caption{The construction process of FinCausal dataset.}
\label{img:FinCausal}
\end{figure}

In this section, we introduce the construction process of the FinCausal dataset, which requires the acquisition of relevant causal knowledge from past financial text materials. This process can be mainly divided into data collection, deduplication, segmentation, and knowledge extraction. The main flowchart is shown in \autoref{img:FinCausal}.

Firstly, we crawled 500,000 research reports from the internet, ranging from 2020 to 2023, including individual stock research, industry analysis, and macro analysis. We used regular matching and the FastText language filter for classification, mainly retaining Chinese A-Share individual stock research and industry analysis. Since individual stock reports rarely describe causal relationships for negative events, we further crawled some news analyses and stock forum comments to enrich the description of individual stock causal knowledge.

After filtering the research report data, we used the MinHash algorithm to perform document-level deduplication on all content. To extract sentences with causal expressions, we meticulously categorized the content of the research reports into seven distinct types, which included ordinary sentences, causal sentences, news-related content, recommendation ratings, investment advice, risk warnings, and researcher information. We then manually annotated a comprehensive dataset comprising 3,000 pieces of data to train a sophisticated BGE+TextCNN classification model. This model was specifically designed to discern and categorize the various types of sentences present in the financial reports, with a particular focus on identifying those that convey causal relationships.


For each piece of extracted knowledge from a research report or comment, we concatenated one sentence before and after it into a paragraph. Subsequently, we aggregated all relevant paragraphs from a report to form a comprehensive context. Utilizing our meticulously chosen LLM, we conducted causal knowledge extraction from these contexts. Through this process, we successfully obtained 200,000 pieces of industry causal knowledge and 2,000 pieces of individual stock causal knowledge, thereby enriching our dataset with valuable insights into the causal relationships within the financial domain. Here are some examples of FinCausal:
\begin{itemize}
\item The company may consider conducting targeted issuance in order to expand its business scale, make capital expenditures, or invest in research and development.
\item During the epidemic, the demand for remote work and online collaboration has increased, driving the development of related software service companies.
\end{itemize}

\clearpage
\section{Result}
\label{sec:a7}


\subsection{Analysis of Direct News Events}
\label{sec:a71}

The examination of news events involves a detailed statistical analysis of the responses generated by various Large Language Models (LLMs) to specific news items. This analysis primarily focuses on the distribution of scores assigned by LLMs to each news event, encompassing key statistical measures such as the mean, variance, highest, and lowest scores. Such an approach is instrumental in assessing the consistency and rationality of LLMs' interpretations of financial news.

To facilitate a comprehensive understanding of these scoring distributions, this section will present violin plots for each news event. Violin plots offer a more nuanced visualization compared to traditional box plots by showing the probability density of the data at different values. This graphical representation will thus provide insights into the spread and skewness of LLMs' ratings across various news events, enabling a deeper analysis of their evaluative patterns and potential biases.



\foreach \i in {0,2,4,...,22} { 
    \begin{figure}[htbp]
    \begin{minipage}{0.48\linewidth}
        \centering
        \includegraphics[width=\linewidth]{fig/news_\i.png}
        \caption{Distribution of the score for news \the\numexpr\i+1\relax ~among 23 large models.}
        \label{img:news_distribution\the\numexpr\i\relax}
    \end{minipage}
    \hfill 
    \begin{minipage}{0.48\linewidth}
        \centering
        \includegraphics[width=\linewidth]{fig/news_\the\numexpr\i+1\relax.png}
        \caption{Distribution of the score for news \the\numexpr\i+2\relax ~among 23 large models.}
        \label{img:news_distribution\the\numexpr\i+1\relax}
    \end{minipage}
    \end{figure}
}


In the analysis of the initial five events, the focus is placed on news items that encompass both positive and negative performance reports, alongside fluctuations in stock prices. This diverse array of news content allows for a multifaceted examination of each Large Language Model's (LLM's) scoring tendencies. Notably, discrepancies in scoring preferences among different LLMs emerge when confronted with this spectrum of financial news.

A systematic statistical analysis is conducted on the scoring outcomes attributed to the positive and negative aspects of these events. This entails a detailed examination of how each LLM assesses the same news piece, shedding light on the variance in their interpretations and the potential implications of their biases. The findings from this analysis are meticulously compiled and presented in \autoref{tab:positive_time}, offering a clear, quantified insight into the LLMs' evaluative patterns across the selected news events.

\begin{table}[htbp]
\centering
\caption{Model Positive Times}
\begin{tabular}{l c}
\hline
\textbf{Model} & \textbf{Positive Times} \\
\hline
GPT-4 & 5 \\
InternLM2-20B & 5 \\
LLaMA2-13B & 5 \\
Qwen-72B & 5 \\
FinQwen & 4 \\
InternLM2-7B & 4 \\
LLaMA2-7B & 4 \\
Qwen-max & 4 \\
Xuanyuan-13B & 4 \\
Baichuan2-13B & 3 \\
Baichuan2-7B & 3 \\
ChatGLM3-Turbo & 3 \\
GLM-4 & 3 \\
Xuanyuan-70B & 3 \\
ChatGLM2-6B & 2 \\
ChatGLM3-6B & 2 \\
Qwen-14B & 2 \\
Qwen-7B & 2 \\
GPT-3.5 & 1 \\
\hline
\end{tabular}
\label{tab:positive_time}
\end{table}

Our analysis involves aggregating the scoring variances observed across all event news for the various Large Language Models (LLMs) under consideration. This comprehensive synthesis not only highlights the diversity in LLM responses but also provides a macroscopic view of their evaluative consistency and potential discrepancies. The aggregated data, which encapsulate the variance in scoring for each news event by different LLMs, are systematically presented in \autoref{tab:news_variance}. This table serves as a pivotal reference point for understanding the range and distribution of LLM evaluations, offering valuable insights into their interpretative frameworks and the reliability of their analyses.

\begin{table}[htbp]
\centering
\caption{Model Variance Comparison}
\begin{tabular}{l l}
\hline
\textbf{Model} & \textbf{Variance} \\
\hline
GLM-4 & 0.59798884 \\
ChatGLM3-6B & 0.707638507 \\
Qwen-72B & 0.784471341 \\
Qwen-7B & 0.788077699 \\
ChatGLM3-Turbo & 1.067120654 \\
Qwen-14B & 1.211226324 \\
Qwen-max & 1.363195393 \\
MiniCPM-2B & 1.409 \\
GPT-4 & 1.909388332 \\
InternLM2-20B & 4.893324616 \\
GPT-3.5 & 5.277003518 \\
Baichuan-13B & 5.998 \\
DISC-FinLLM & 6.096 \\
Baichuan2-13B & 6.157081681 \\
LLaMA2-13B & 6.881628177 \\
InternLM2-7B & 7.466014898 \\
FinQwen & 9.363445905 \\
ChatGLM2-6B & 10.03005785 \\
LLaMA2-7B & 10.61274671 \\
Xuanyuan-70B & 10.83743438 \\
Xuanyuan2-6B & 13.9988 \\
Xuanyuan-13B & 19.18007393 \\
Baichuan2-7B & 28.10579705 \\
\hline
\end{tabular}
\label{tab:news_variance}
\end{table}


Upon examining the inherent biases within individual models, our analysis proceeds to consolidate the findings from each Large Language Model (LLM) to explore their collective or differential biases towards various industries. This step is crucial for understanding not only the predispositions of individual models but also for discerning any overarching trends or anomalies in their assessments of industry-related news events. By aggregating these results, we aim to delineate the extent to which these models exhibit preferential or adverse biases towards certain industry, thereby shedding light on the potential influence of these biases on the models' analytical outputs and reliability. The synthesis of this comprehensive analysis provides a nuanced understanding of model behavior in the context of industry-specific evaluations.

\begin{figure}[htbp]
\begin{minipage}{0.48\linewidth}
 \centering
 \includegraphics[width=\linewidth]{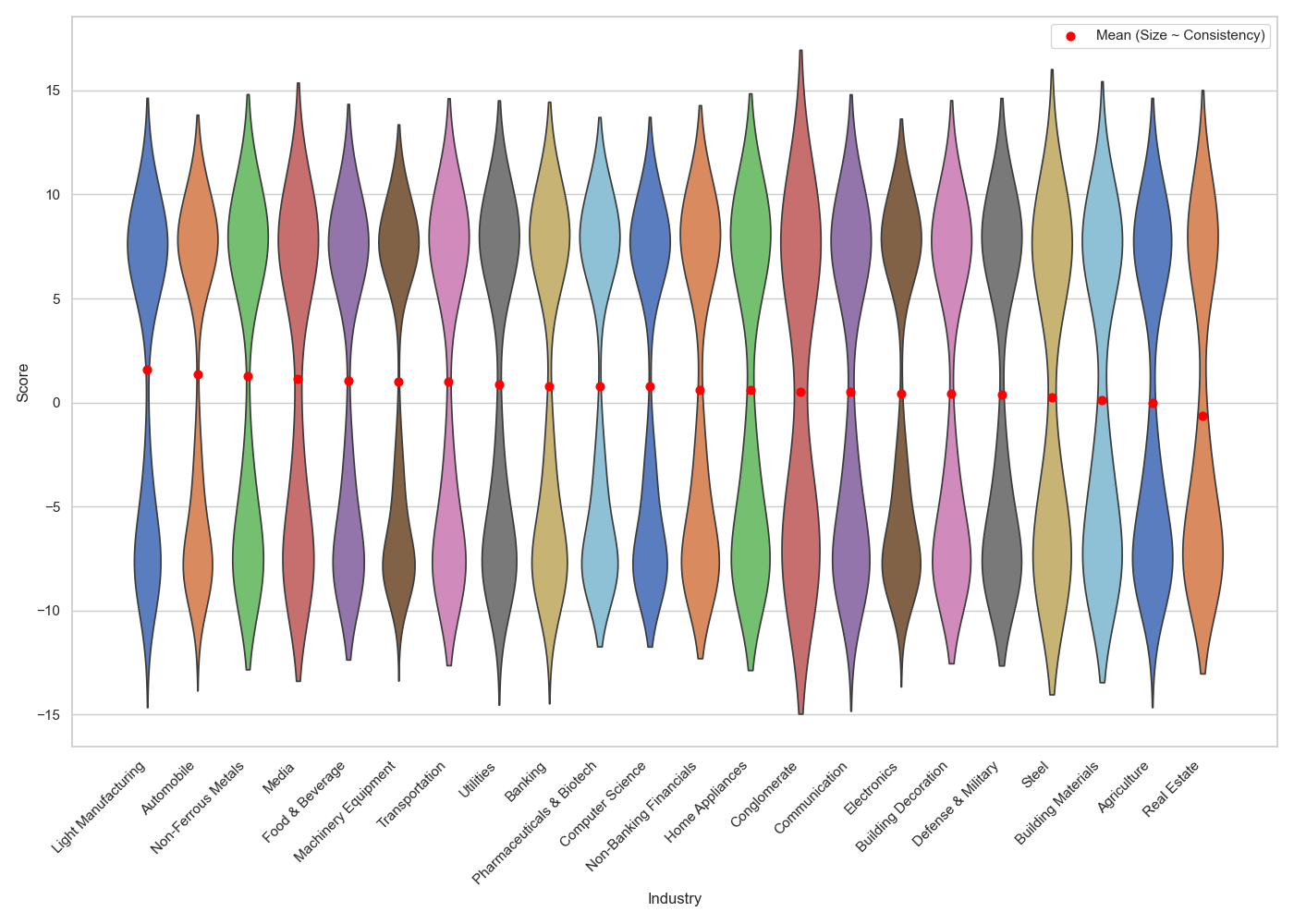}
 \caption{Distribution of the industry scores of Baichuan2-7B.}
 \label{img:industry_fig1}
\end{minipage}
\hfill 
\begin{minipage}{0.48\linewidth}
 \centering
 \includegraphics[width=\linewidth]{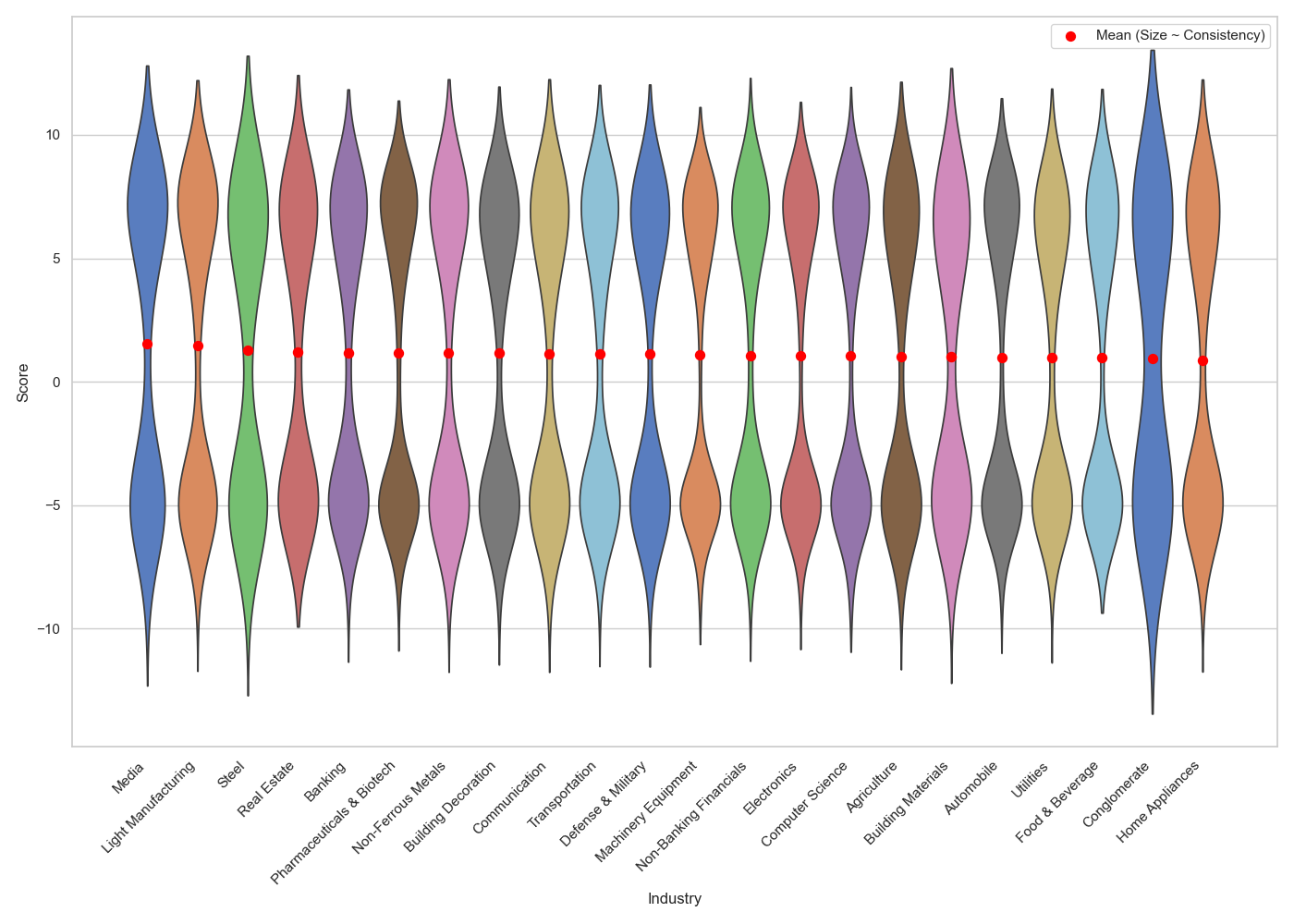}
 \caption{Distribution of the industry scores of Baichuan2-13B.}
 \label{img:industry_fig2}
\end{minipage}
\end{figure}

\begin{figure}[htbp]
\begin{minipage}{0.48\linewidth}
 \centering
 \includegraphics[width=\linewidth]{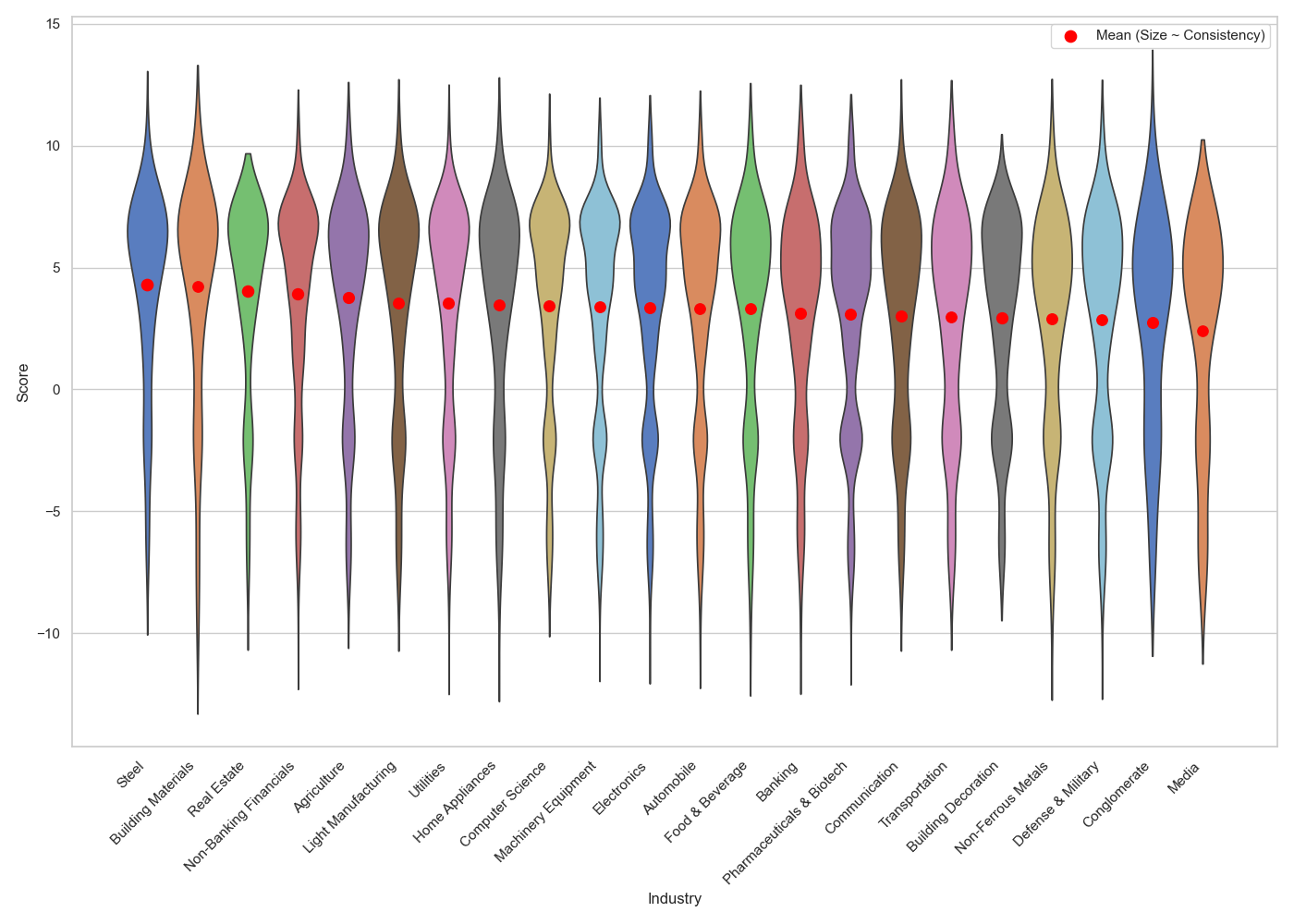}
 \caption{Distribution of the industry scores of ChatGLM2-6B.}
 \label{img:industry_fig3}
\end{minipage}
\hfill 
\begin{minipage}{0.48\linewidth}
 \centering
 \includegraphics[width=\linewidth]{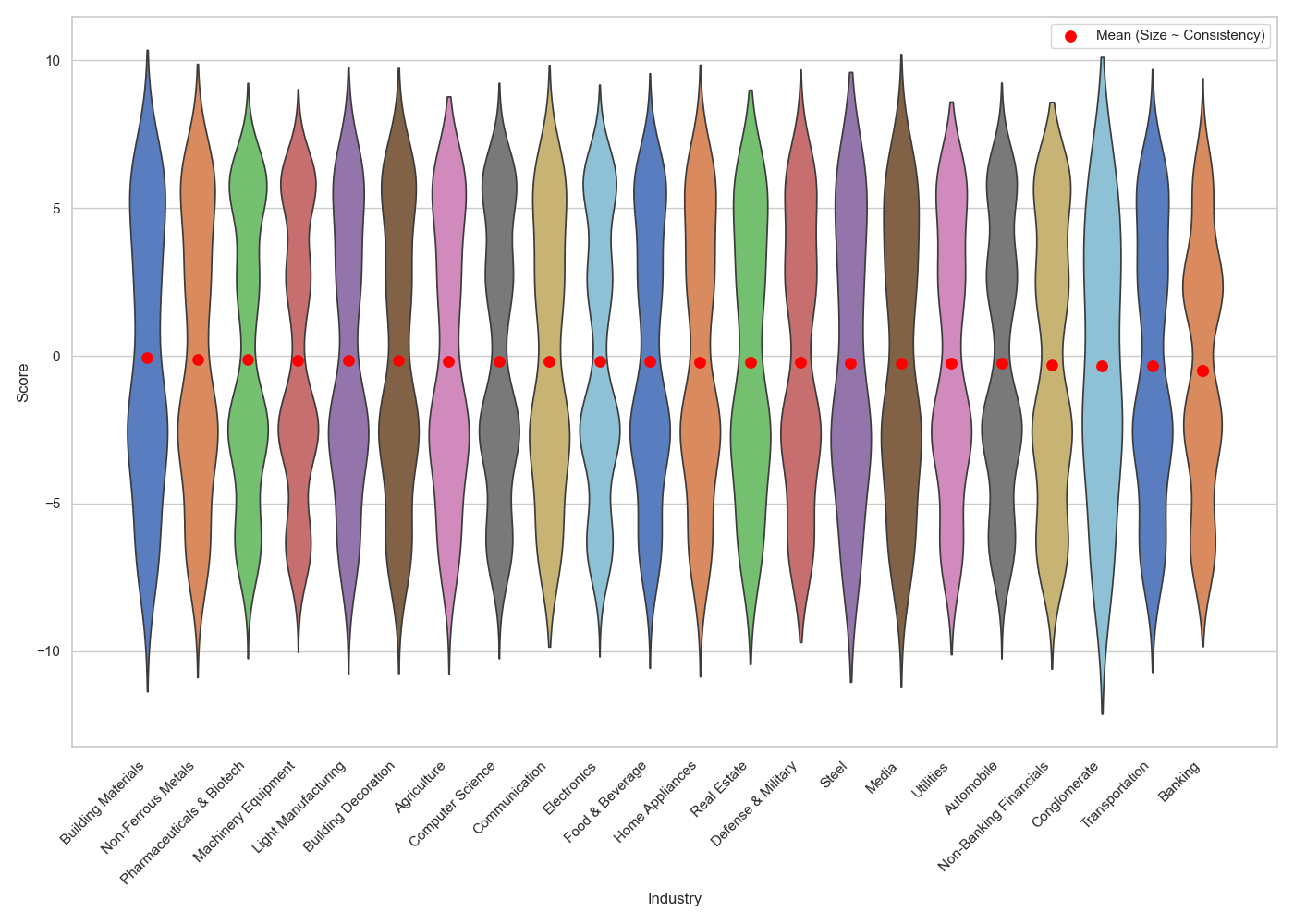}
 \caption{Distribution of the industry scores of ChatGLM3-6B.}
 \label{img:industry_fig4}
\end{minipage}
\end{figure}

\begin{figure}[htbp]
\begin{minipage}{0.48\linewidth}
 \centering
 \includegraphics[width=\linewidth]{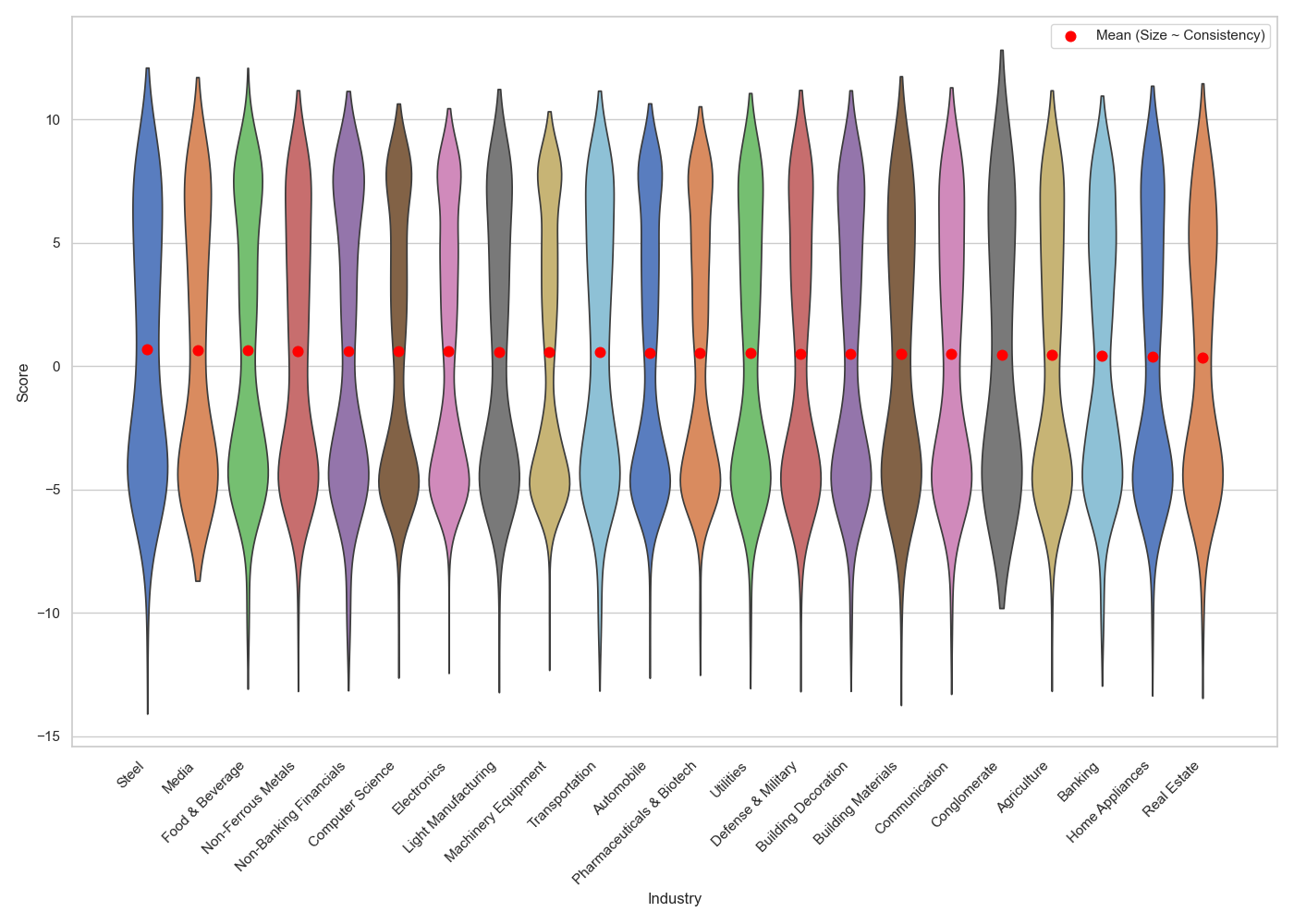}
 \caption{Distribution of the industry scores of ChatGLM3-Turbo.}
 \label{img:industry_fig5}
\end{minipage}
\hfill
\begin{minipage}{0.48\linewidth}
 \centering
 \includegraphics[width=\linewidth]{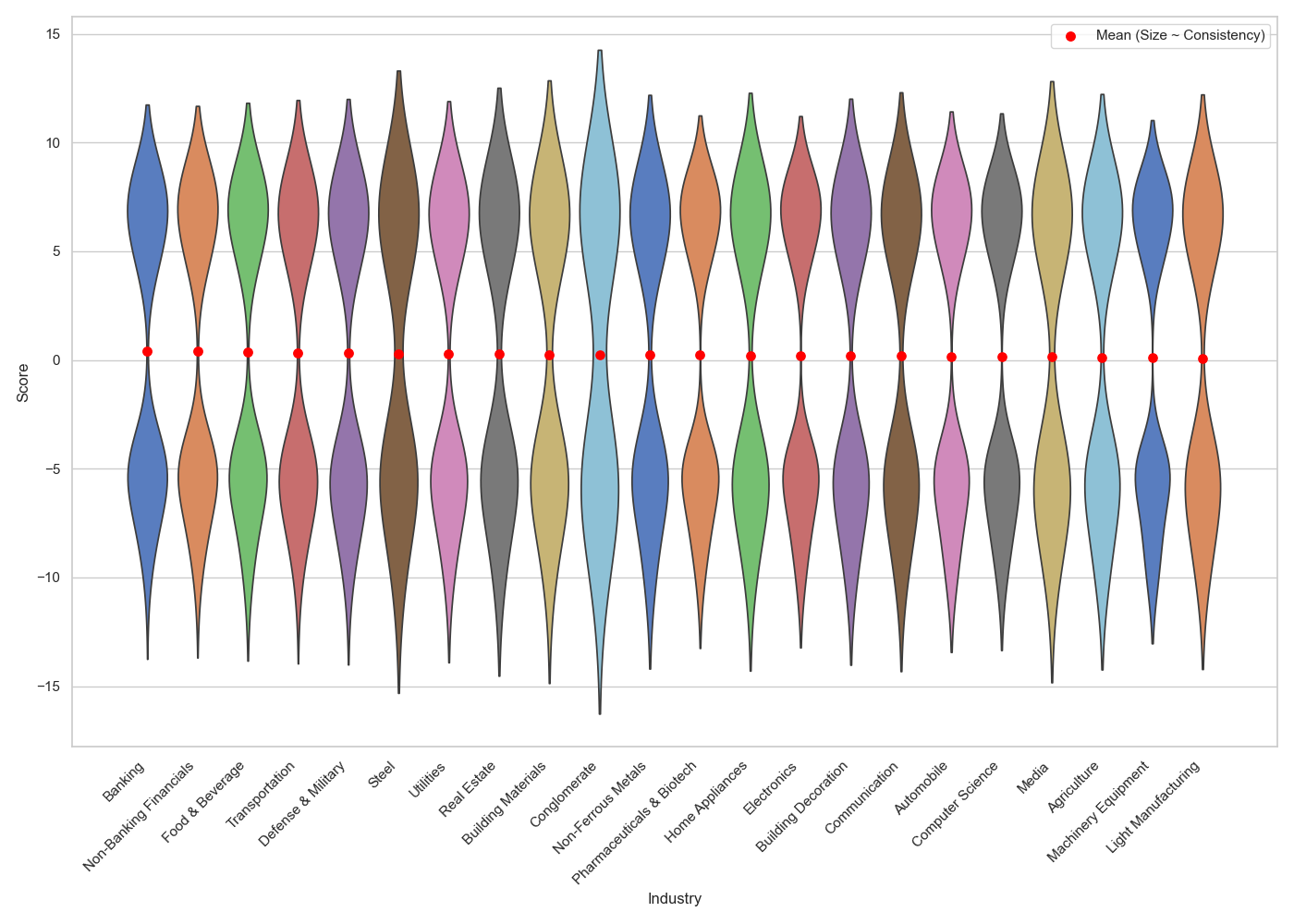}
 \caption{Distribution of the industry scores of GLM-4.}
 \label{img:industry_fig6}
\end{minipage}
\end{figure}

\begin{figure}[htbp]
\begin{minipage}{0.48\linewidth}
 \centering
 \includegraphics[width=\linewidth]{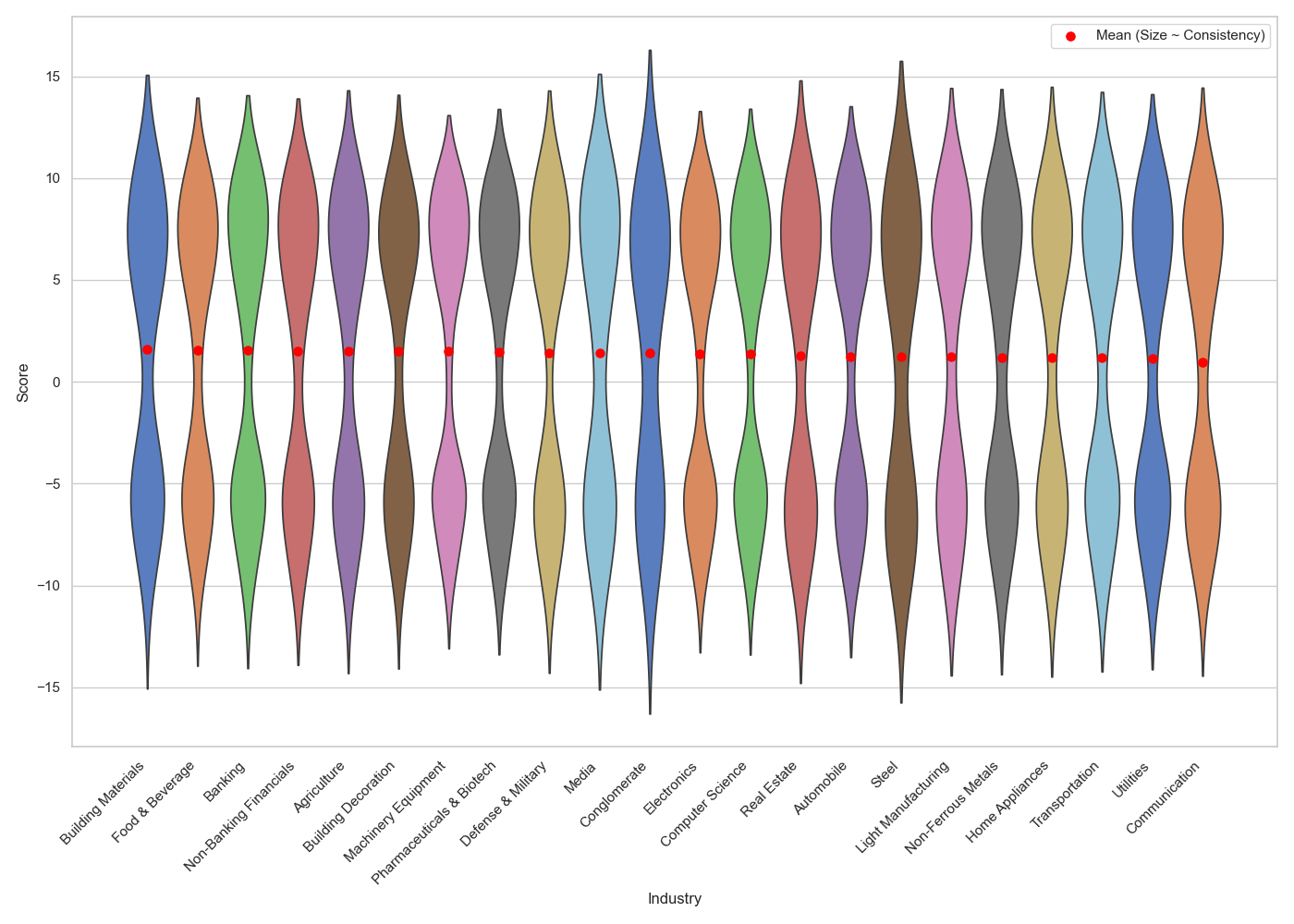}
 \caption{Distribution of the industry scores of InternLM2-7B.}
 \label{img:industry_fig7}
\end{minipage}
\hfill
\begin{minipage}{0.48\linewidth}
 \centering
 \includegraphics[width=\linewidth]{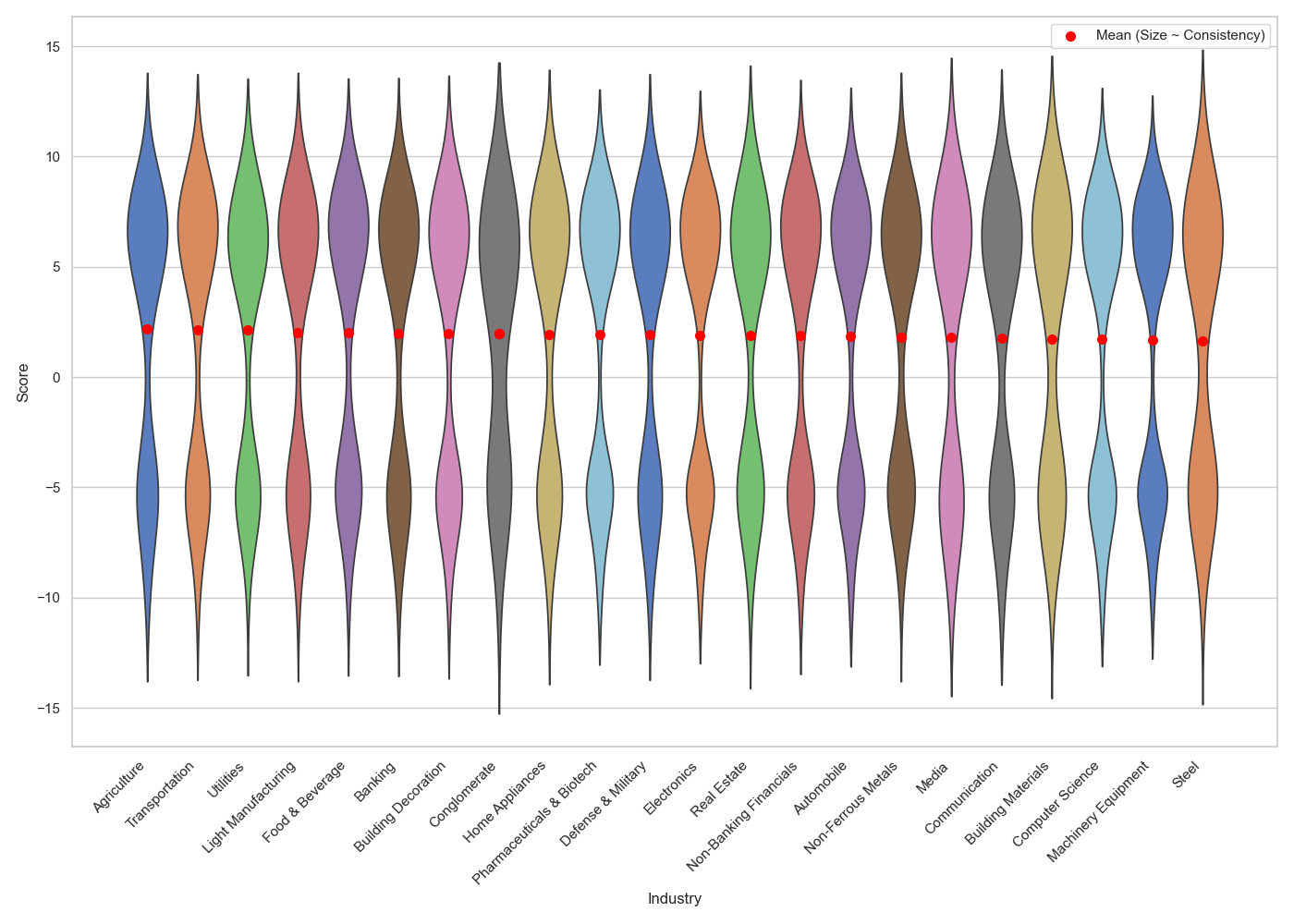}
 \caption{Distribution of the industry scores of InternLM2-20B.}
 \label{img:industry_fig8}
\end{minipage}
\end{figure}

\begin{figure}[htbp]
\begin{minipage}{0.48\linewidth}
 \centering
 \includegraphics[width=\linewidth]{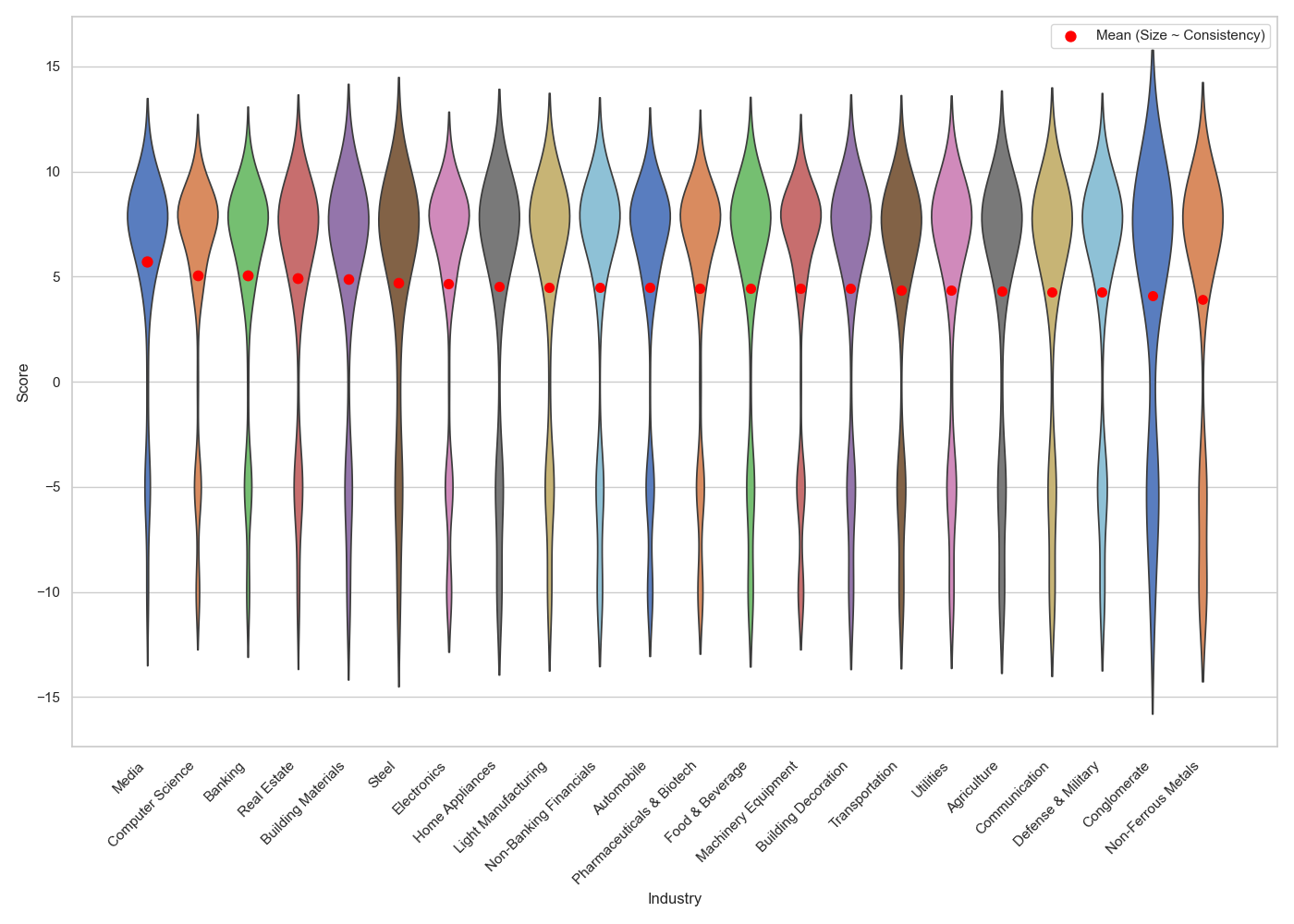}
 \caption{Distribution of the industry scores of LLaMA2-7B.}
 \label{img:industry_fig9}
\end{minipage}
\hfill
\begin{minipage}{0.48\linewidth}
 \centering
 \includegraphics[width=\linewidth]{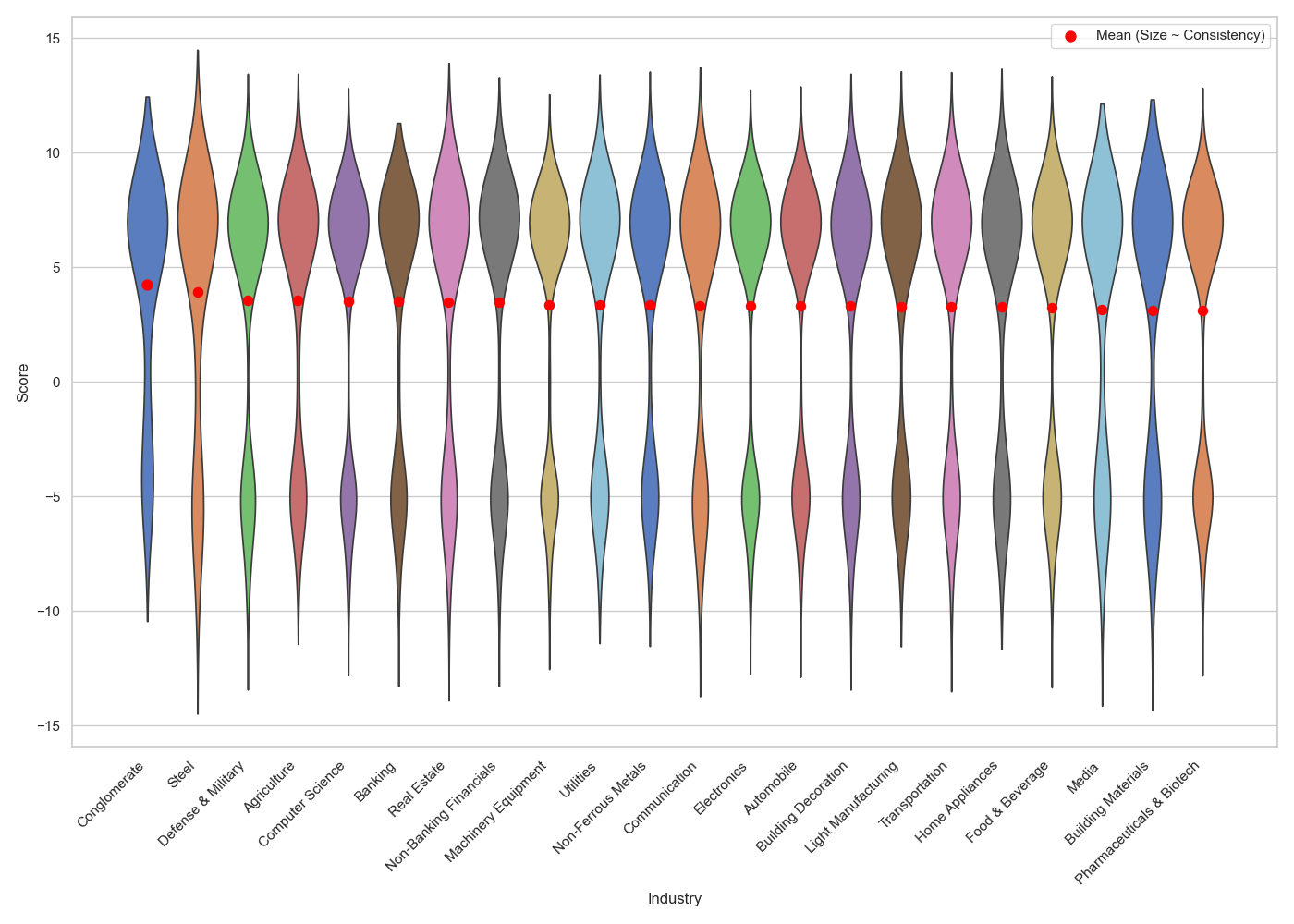}
 \caption{Distribution of the industry scores of LLaMA2-13B.}
 \label{img:industry_fig10}
\end{minipage}
\end{figure}

\begin{figure}[htbp]
\begin{minipage}{0.48\linewidth}
 \centering
 \includegraphics[width=\linewidth]{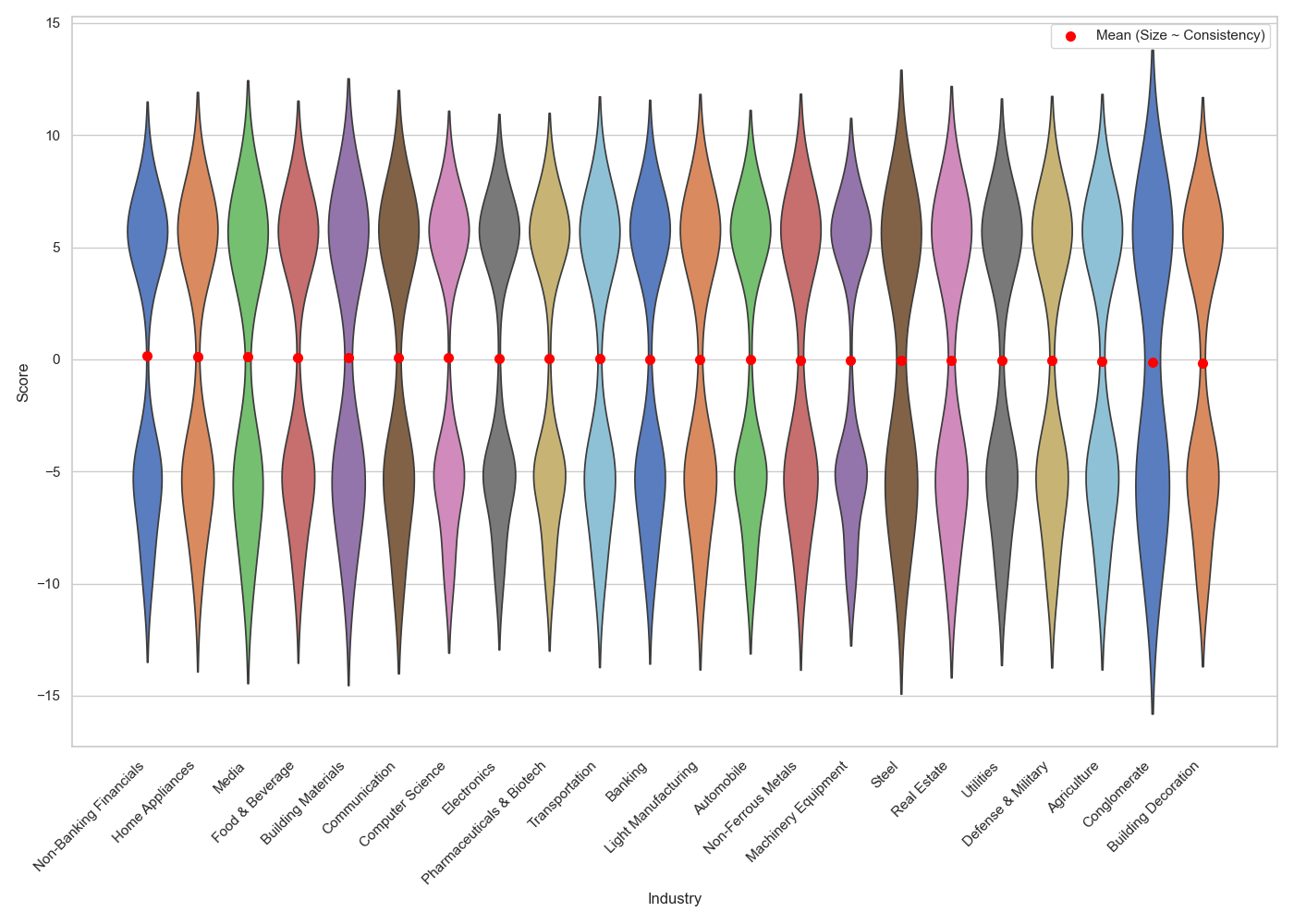}
 \caption{Distribution of the industry scores of Qwen-7B.}
 \label{img:industry_fig11}
\end{minipage}
\hfill
\begin{minipage}{0.48\linewidth}
 \centering
 \includegraphics[width=\linewidth]{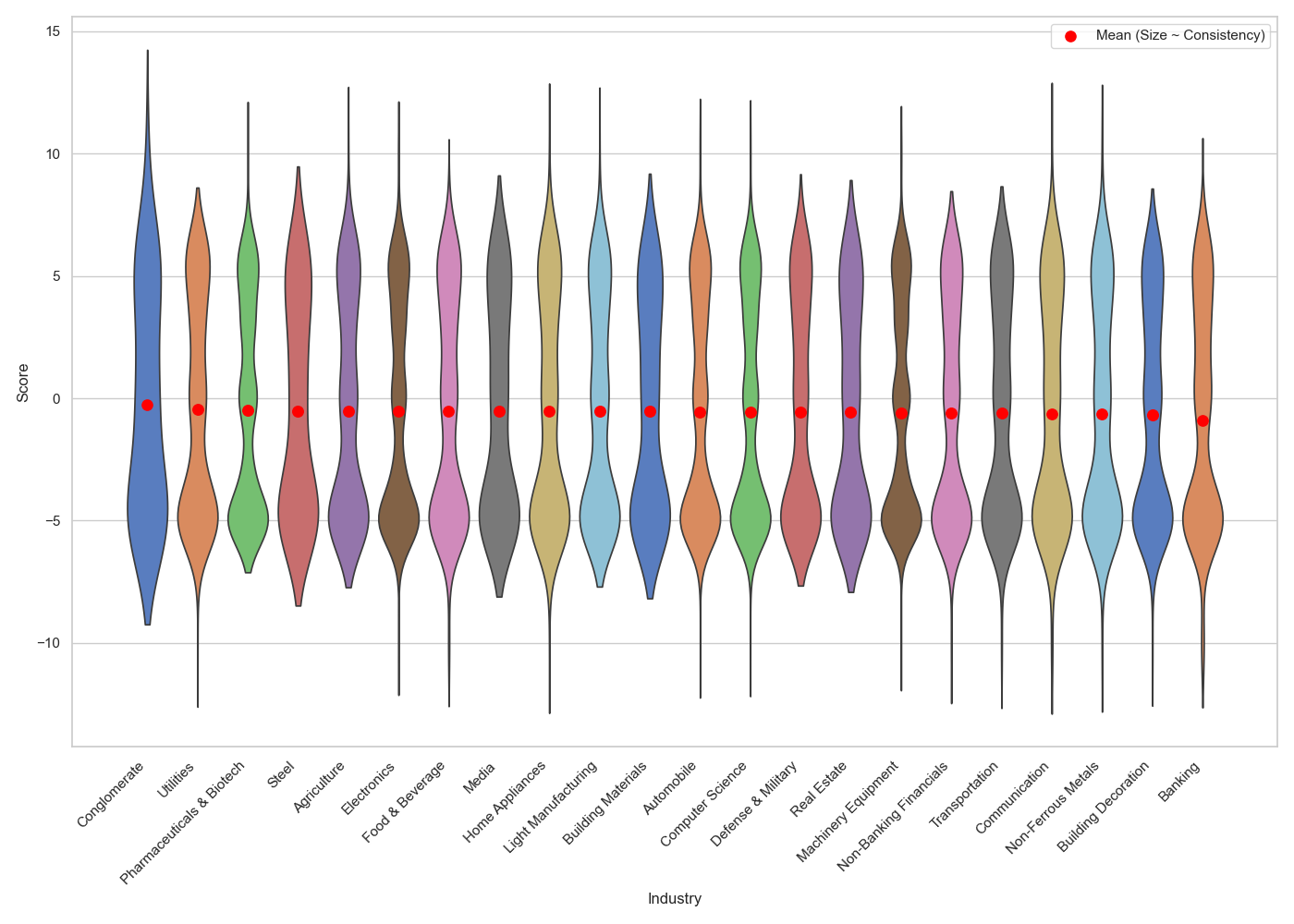}
 \caption{Distribution of the industry scores of Qwen-14B.}
 \label{img:industry_fig12}
\end{minipage}
\end{figure}

\begin{figure}[htbp]
\begin{minipage}{0.48\linewidth}
 \centering
 \includegraphics[width=\linewidth]{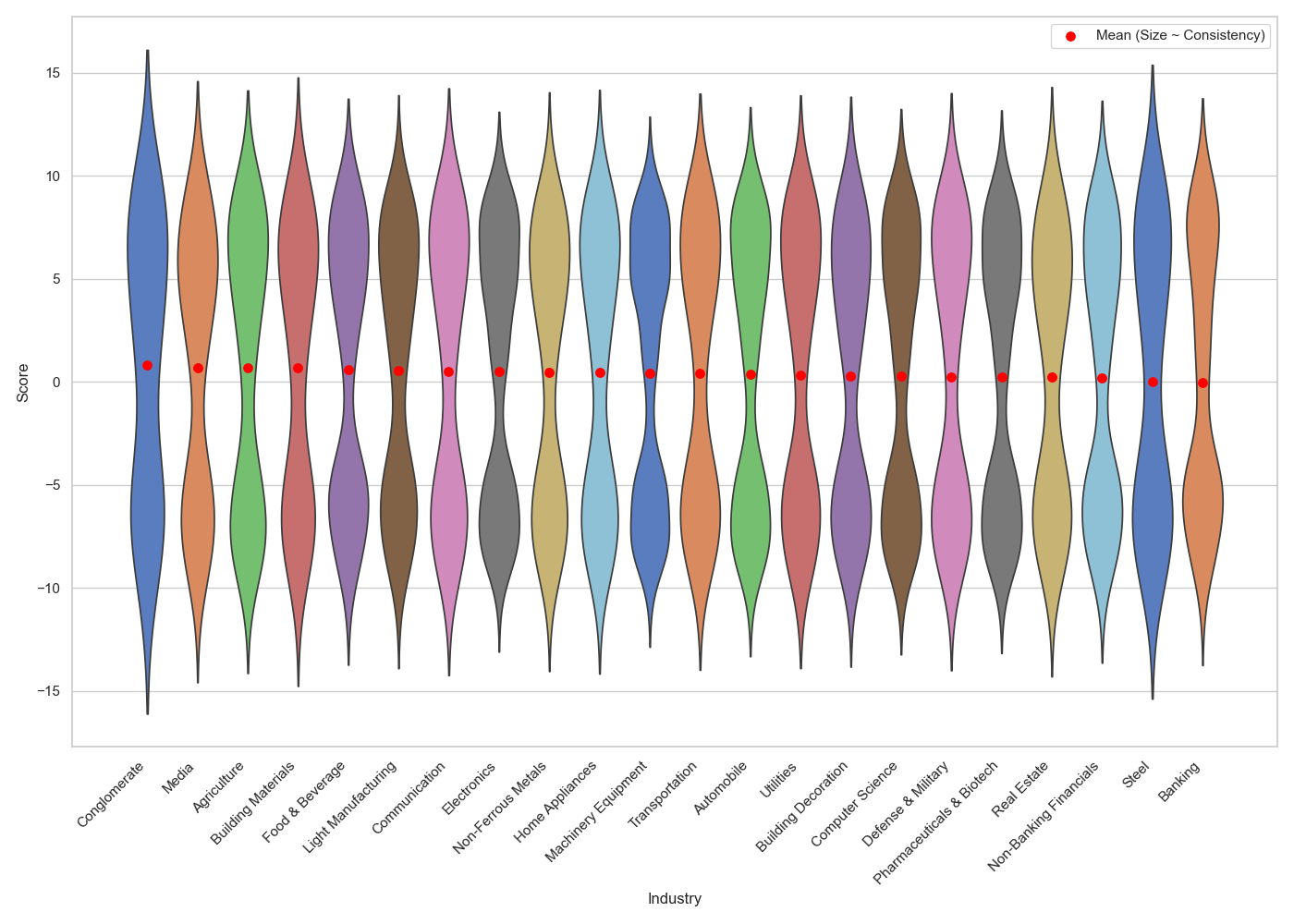}
 \caption{Distribution of the industry scores of FinQwen.}
 \label{img:industry_fig13}
\end{minipage}
\hfill
\begin{minipage}{0.48\linewidth}
 \centering
 \includegraphics[width=\linewidth]{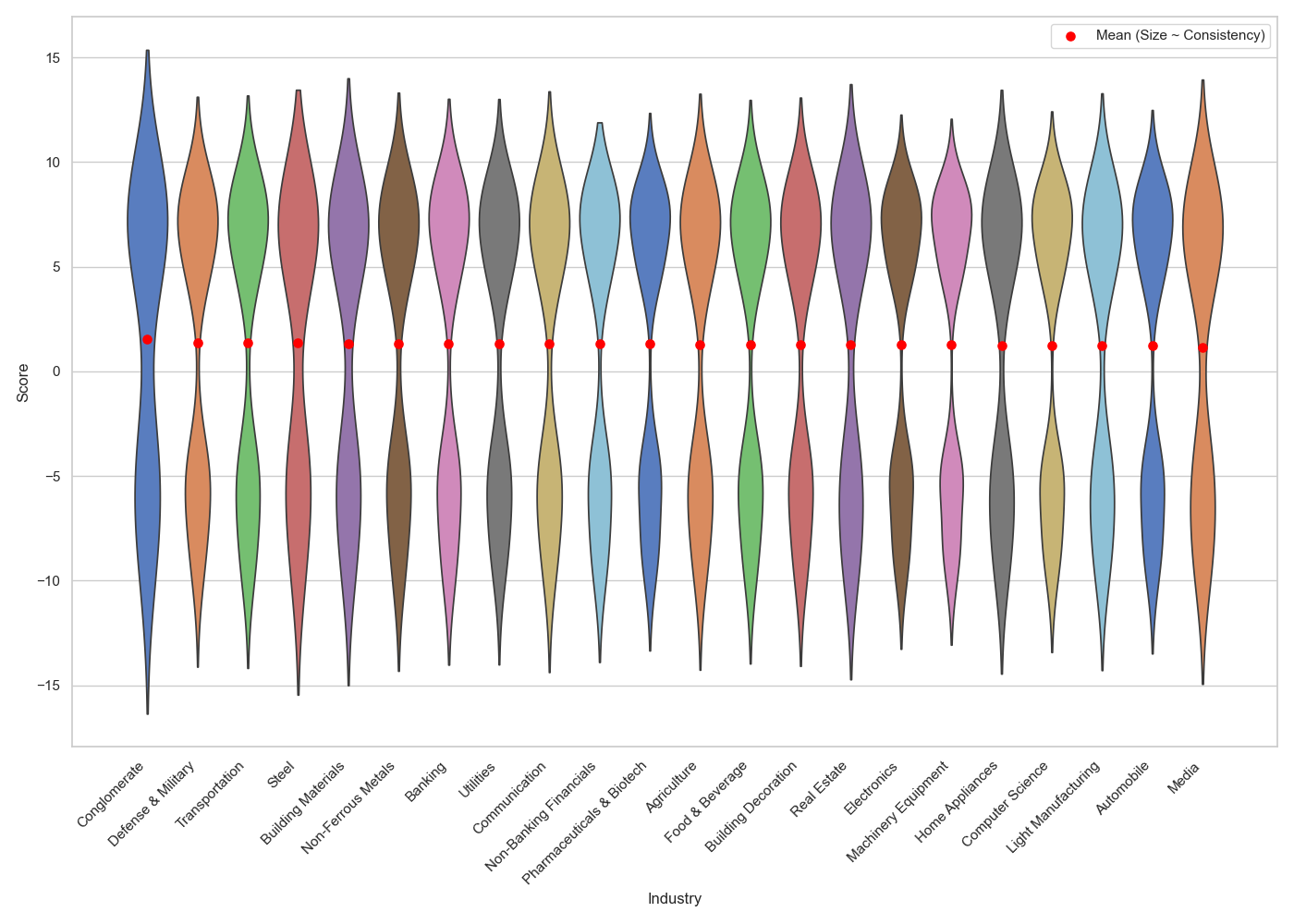}
 \caption{Distribution of the industry scores of Qwen-72B.}
 \label{img:industry_fig14}
\end{minipage}
\end{figure}

\begin{figure}[htbp]
\begin{minipage}{0.48\linewidth}
 \centering
 \includegraphics[width=\linewidth]{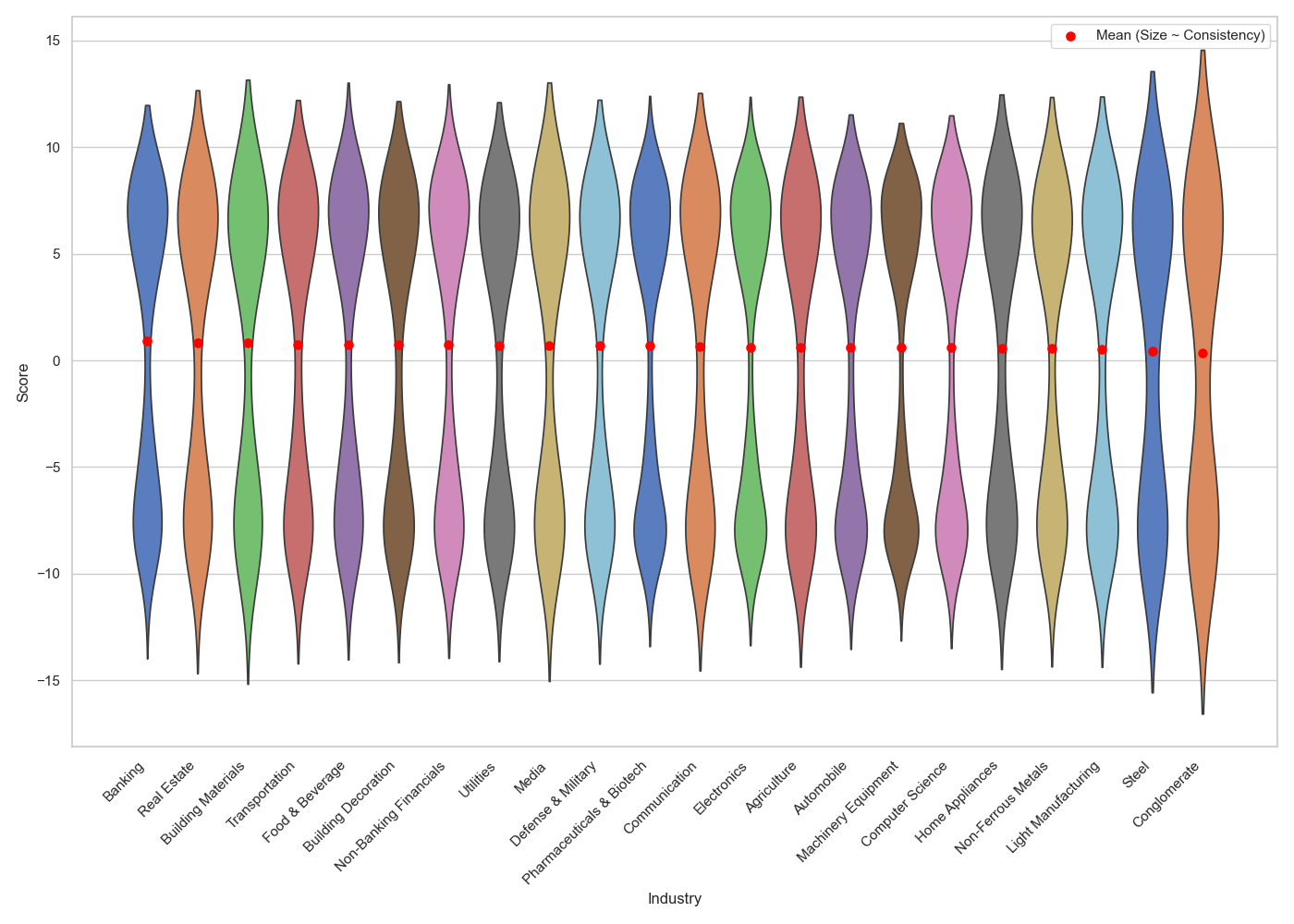}
 \caption{Distribution of the industry scores of Qwen-max.}
 \label{img:industry_fig15}
\end{minipage}
\hfill
\begin{minipage}{0.48\linewidth}
 \centering
 \includegraphics[width=\linewidth]{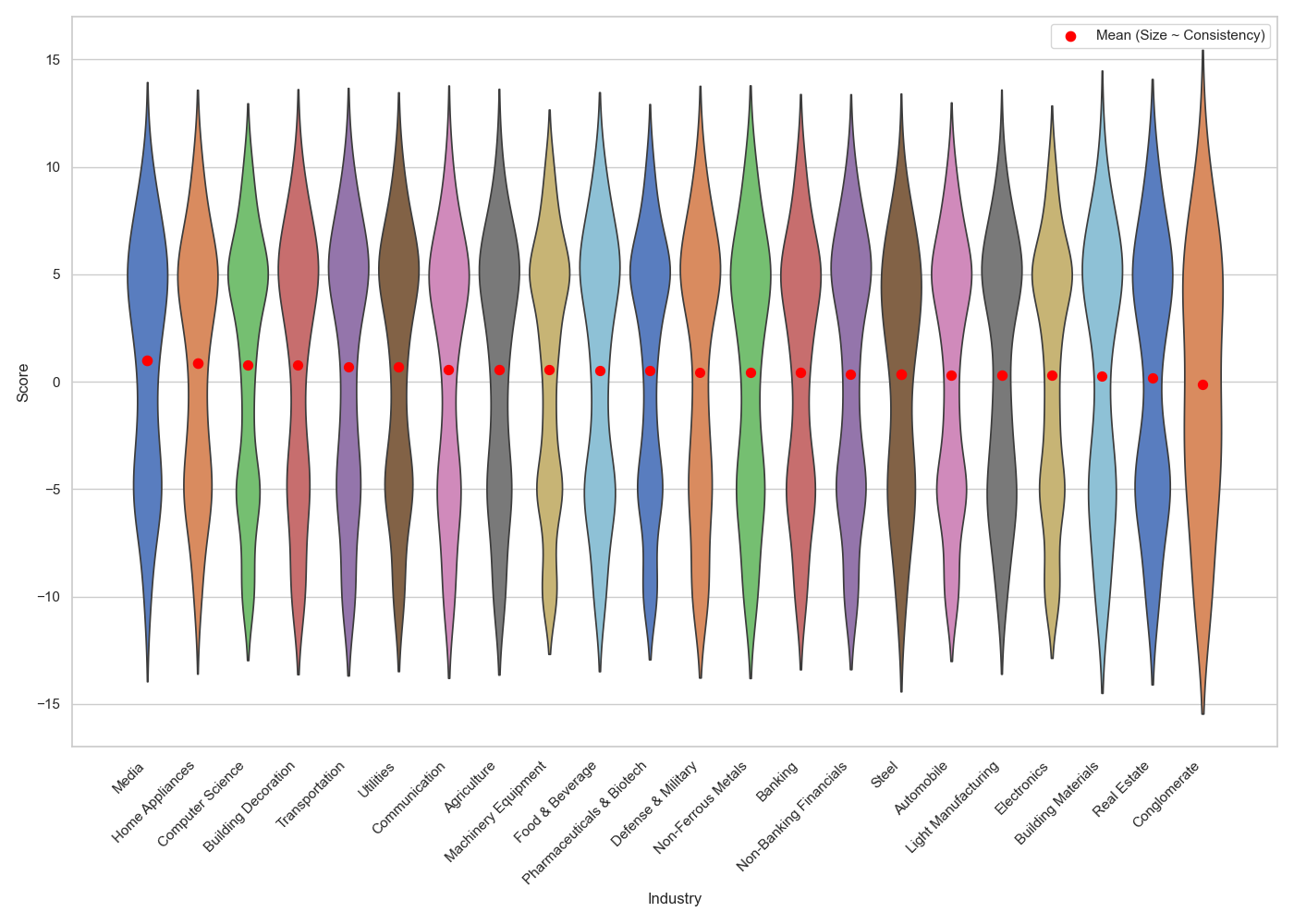}
 \caption{Distribution of the industry scores of Xuanyuan-13B.}
 \label{img:industry_fig16}
\end{minipage}
\end{figure}

\begin{figure}[htbp]
\begin{minipage}{0.48\linewidth}
 \centering
 \includegraphics[width=\linewidth]{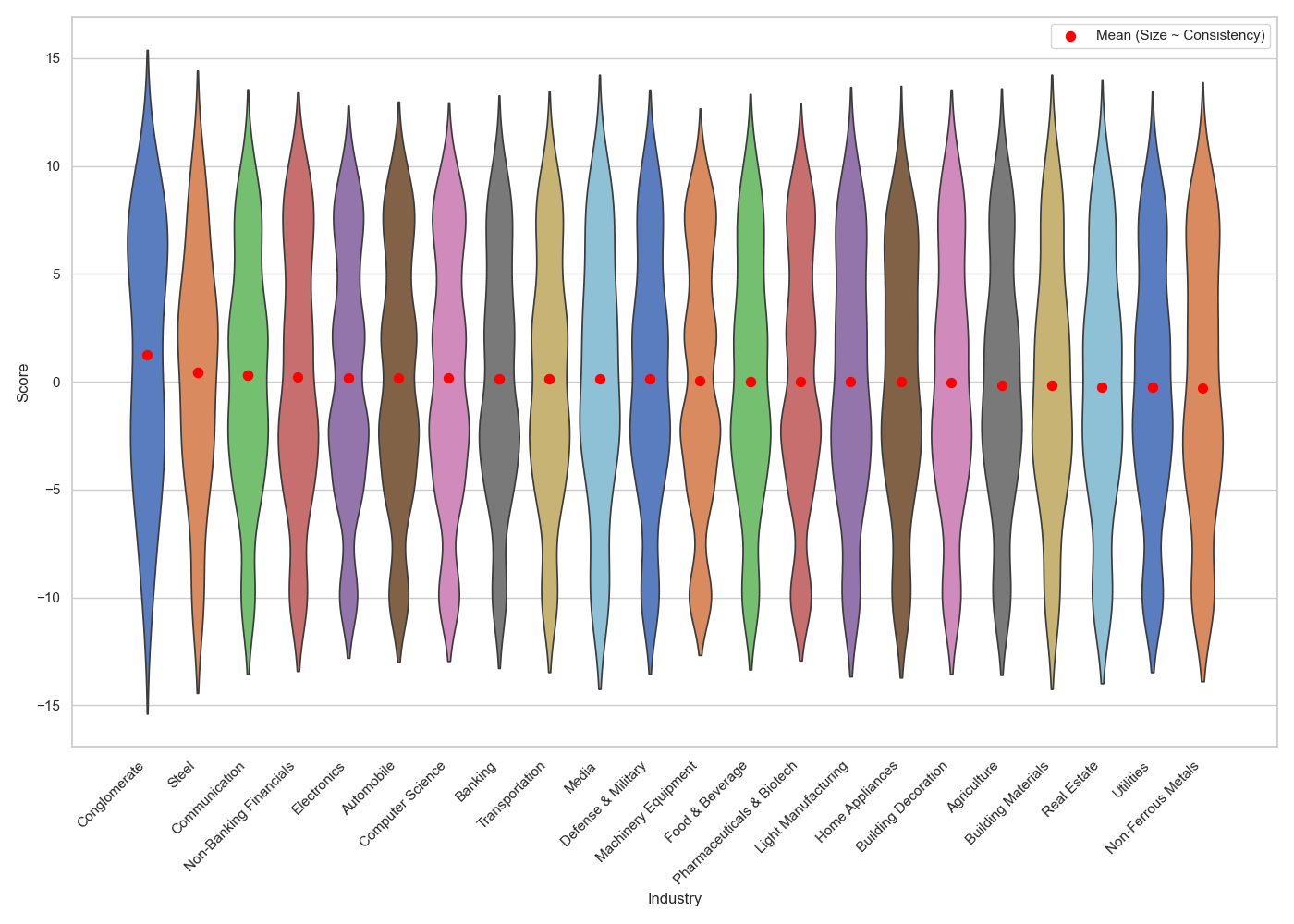}
 \caption{Distribution of the industry scores of Xuanyuan-70B.}
 \label{img:industry_fig17}
\end{minipage}
\hfill
\begin{minipage}{0.48\linewidth}
 \centering
 \includegraphics[width=\linewidth]{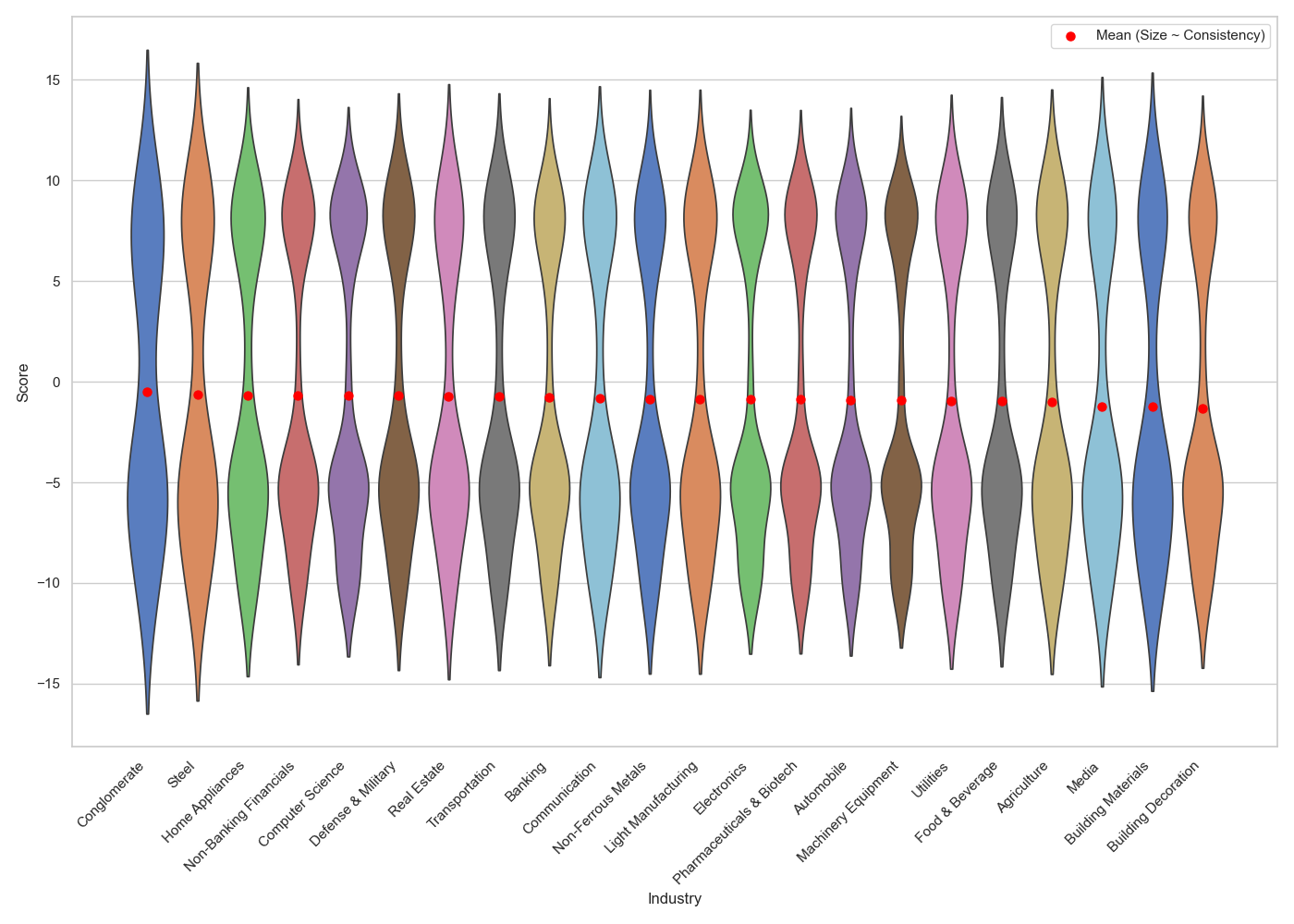}
 \caption{Distribution of the industry scores of GPT-3.5.}
 \label{img:industry_fig18}
\end{minipage}
\end{figure}

\begin{figure}[htbp]
\centering
\includegraphics[width=0.48\linewidth]{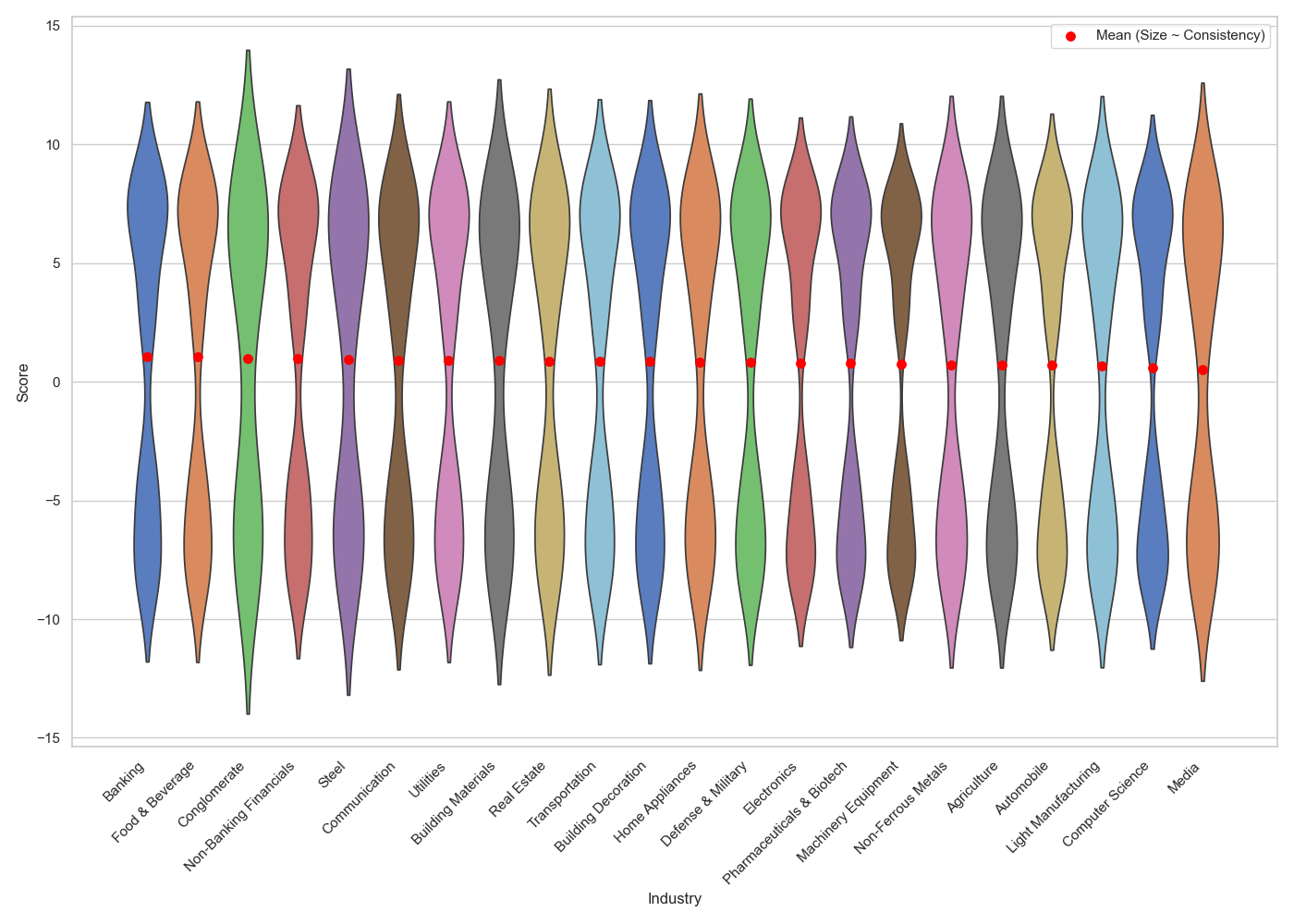}
\caption{Distribution of the industry scores of GPT-4.}
\label{img:industry_fig19}
\end{figure}

\begin{figure}[htbp]
\centerline{\includegraphics[width=0.9\linewidth]{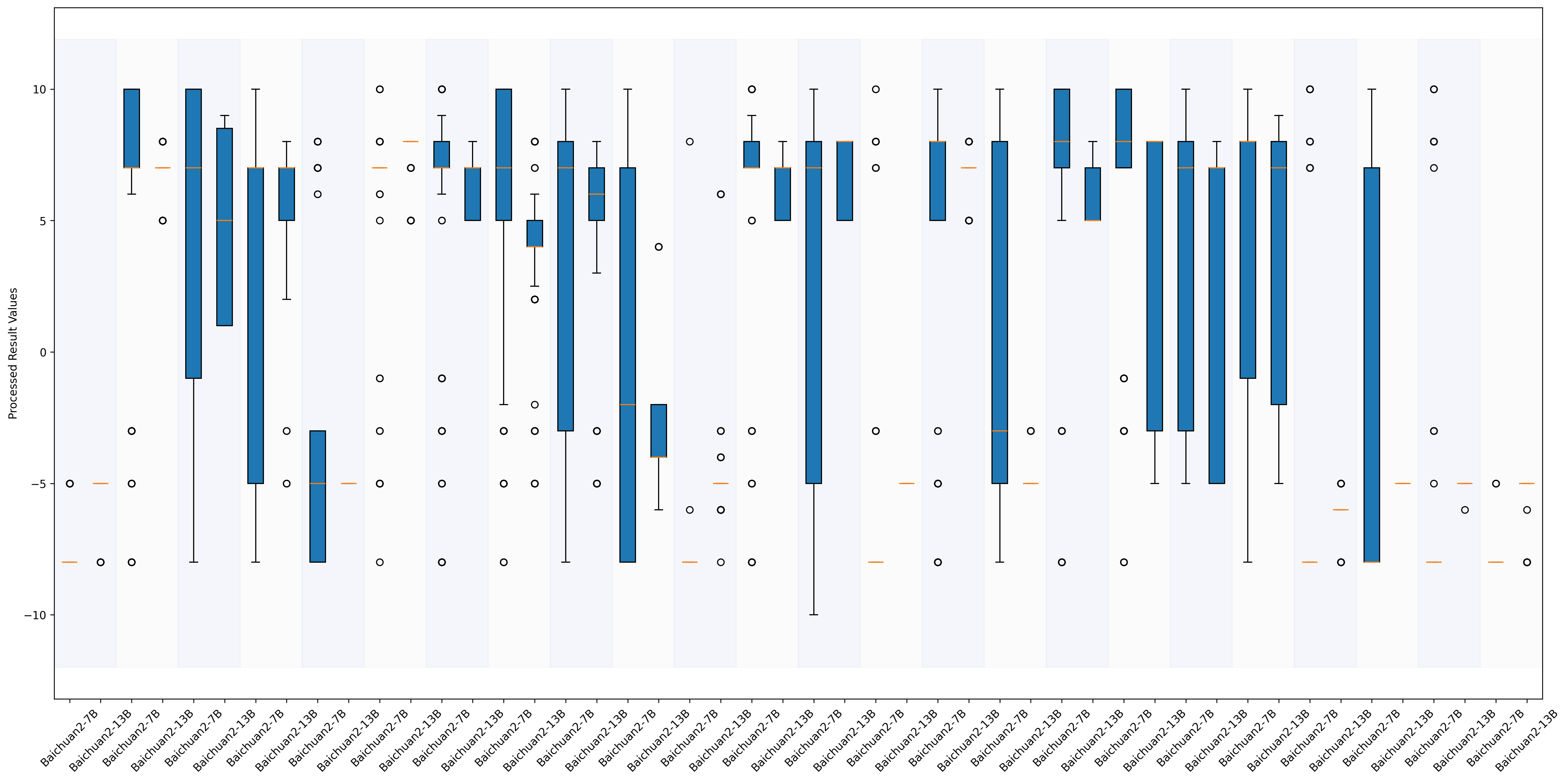}}
\caption{Box line comparison charts of Baichuan family.}
\label{img:baichuan_series}
\end{figure}
\begin{figure}[htbp]
\centerline{\includegraphics[width=0.9\linewidth]{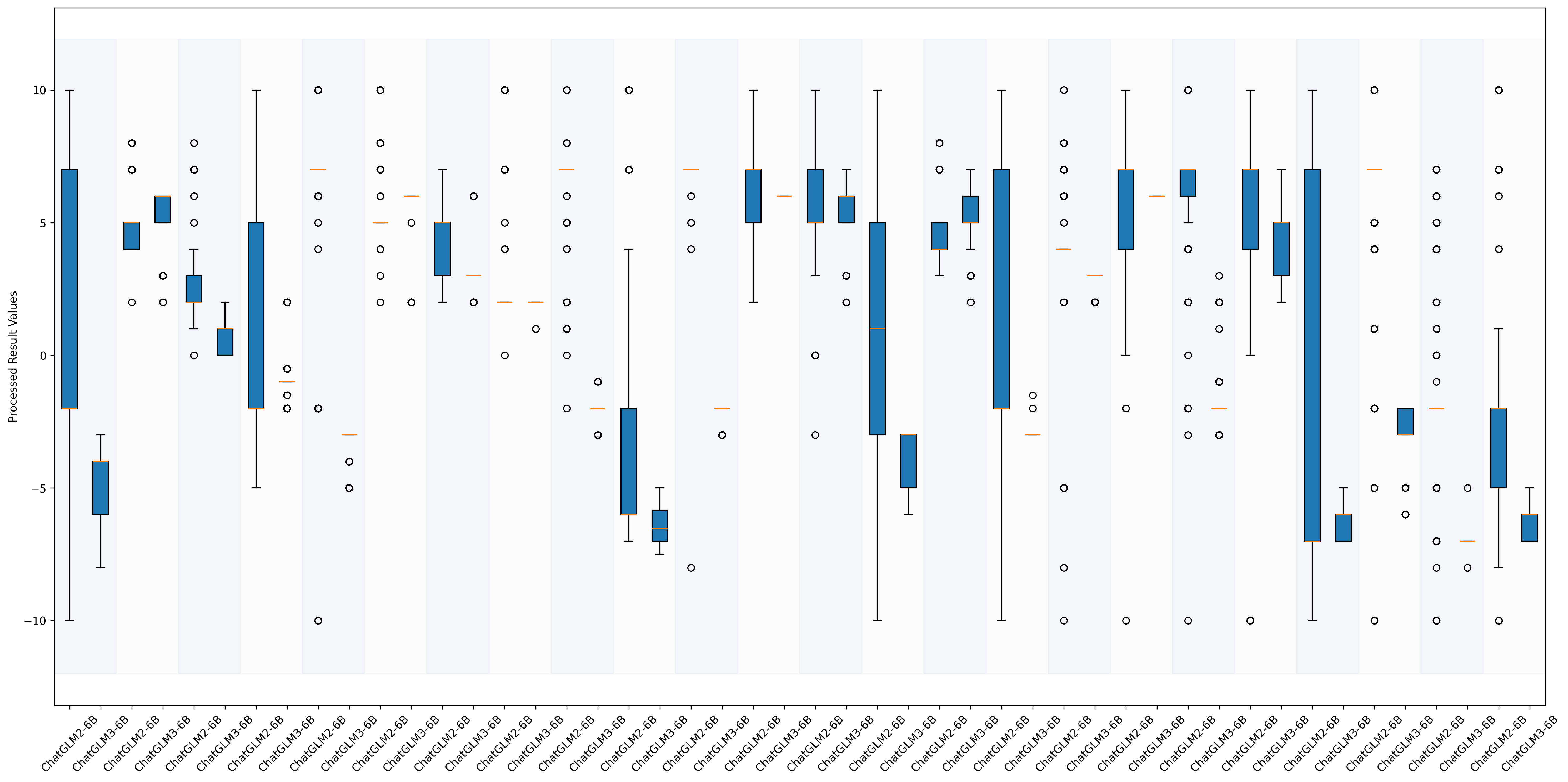}}
\caption{Box line comparison charts of GLM family.}
\label{img:glm_series}
\end{figure}
\begin{figure}[htbp]
\centerline{\includegraphics[width=0.9\linewidth]{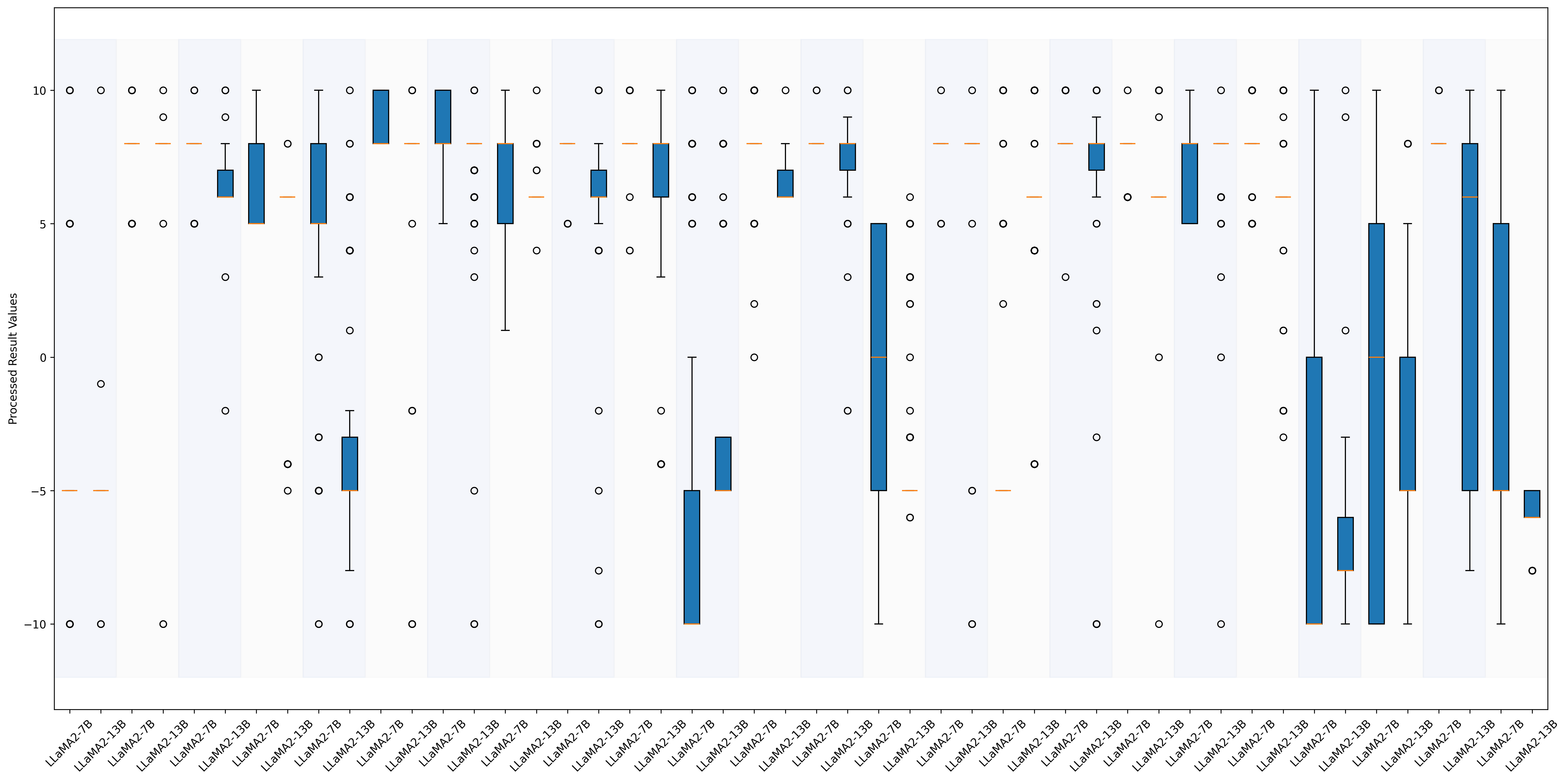}}
\caption{Box line comparison charts of LLaMA family.}
\label{img:LLaMa_series}
\end{figure}

In parallel with the examination of model biases, our study also delves into the temporal evolution of Large Language Models within distinct family, attributing changes to factors such as model size or software updates. To this end, we employ box line comparison charts as a visual tool to elucidate the developmental trajectories of models within the Baichuan, GLM, and LLaMA family. These charts serve to highlight variations in model performance or bias over time, providing a clear visual representation of progression or shifts in model behavior. The comparative analyses for the Baichuan family, GLM family, and LLaMA family are depicted in \autoref{img:baichuan_series}, \autoref{img:glm_series}, and \autoref{img:LLaMa_series}, respectively. Through these visual comparisons, we aim to offer insights into how advancements or modifications in model architecture and capabilities influence their analytical outcomes and biases.

\subsection{Analysis of COT News}
\label{sec:a72}
To delve deeper into the underlying factors contributing to potential irrationalities in Large Language Models (LLMs), our investigation extends to the analysis of reasoning outcomes facilitated by cognitive connections. This approach is predicated on the hypothesis that the manner in which LLMs forge and utilize cognitive links during the reasoning process may shed light on their logical inconsistencies or biases. For this purpose, we have meticulously selected LLMs that have demonstrated the highest, second highest, and lowest levels of performance in response to direct prompts. This selection criterion ensures a comprehensive overview, encompassing a broad spectrum of reasoning capabilities within LLMs.

The focus of this analysis is on the 'slow thinking' aspect of model reasoning, where deliberate and methodical processing is emphasized. By examining the variance in reasoning outcomes among these models, we aim to identify patterns or anomalies that might indicate a propensity for irrational decision-making. The results of this analysis, highlighting the variance in cognitive reasoning among the selected LLMs, are systematically presented in \autoref{tab:cot_variance}. Through this examination, we seek to uncover the intricacies of cognitive processing in LLMs and their implications for model reliability and rationality.

\begin{table}[htbp]
\centering
\caption{Model Variance Comparison after COT}
\begin{tabular}{l l l}
\hline
\textbf{Model} & \textbf{Direct} & \textbf{COT} \\
\hline
GLM-4 & 0.597988840 & 5.381799977 \\
Qwen-7B & 0.788077699 & 7.685484073 \\
ChatGLM3-Turbo & 1.067120654 & 5.670563704 \\
MiniCPM-2B & 1.409476674 & 7.038961848 \\
Xuanyuan2-6B & 13.99883011 & 6.668694967 \\
Xuanyuan-13B & 19.18007393 & 17.83545943 \\
Baichuan2-7B & 28.10579705 & 12.65975644 \\
\hline
\end{tabular}
\label{tab:cot_variance}
\end{table}

At the same time, we further analyzed the new5 with significant differences in ratings among different LLMs, and used the keyword detection method Bertopic to cluster and analyze the reasoning results of the models. Before clustering, the model scores and specific information of the company were removed from the reasoning results. The inference decibel of each model is clustered into 10 categories, and the distribution of scores for each category is as follows.

\begin{figure}[htbp]
\begin{minipage}{0.48\linewidth}
 \centering
    \includegraphics[width=1\linewidth]{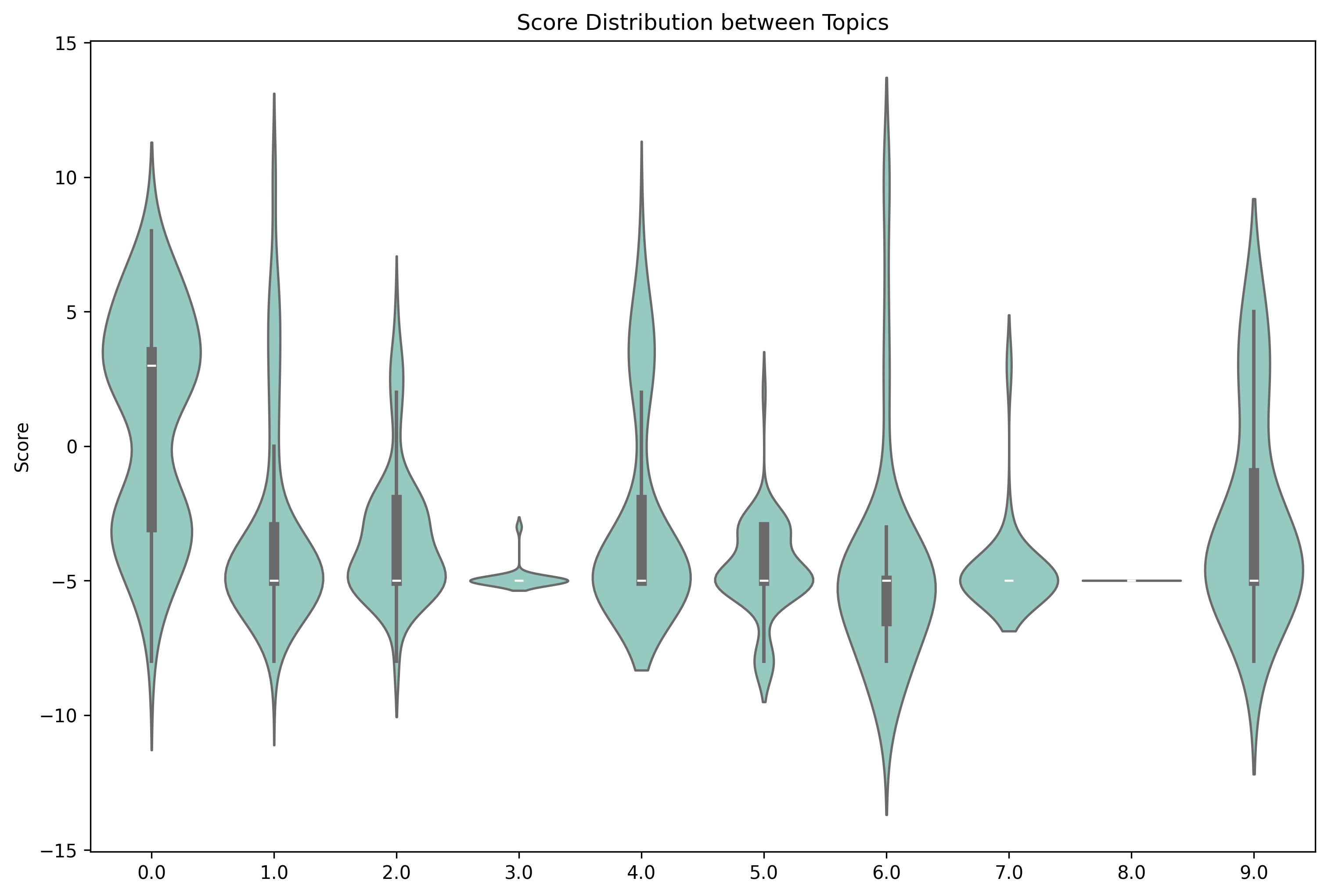}
    \caption{Distribution of cluster scores of Baichuan2-7B.}
    \label{img:baichuan_reasoning}
\end{minipage}
\hfill
\begin{minipage}{0.48\linewidth}
 \centering
 \includegraphics[width=1\linewidth]{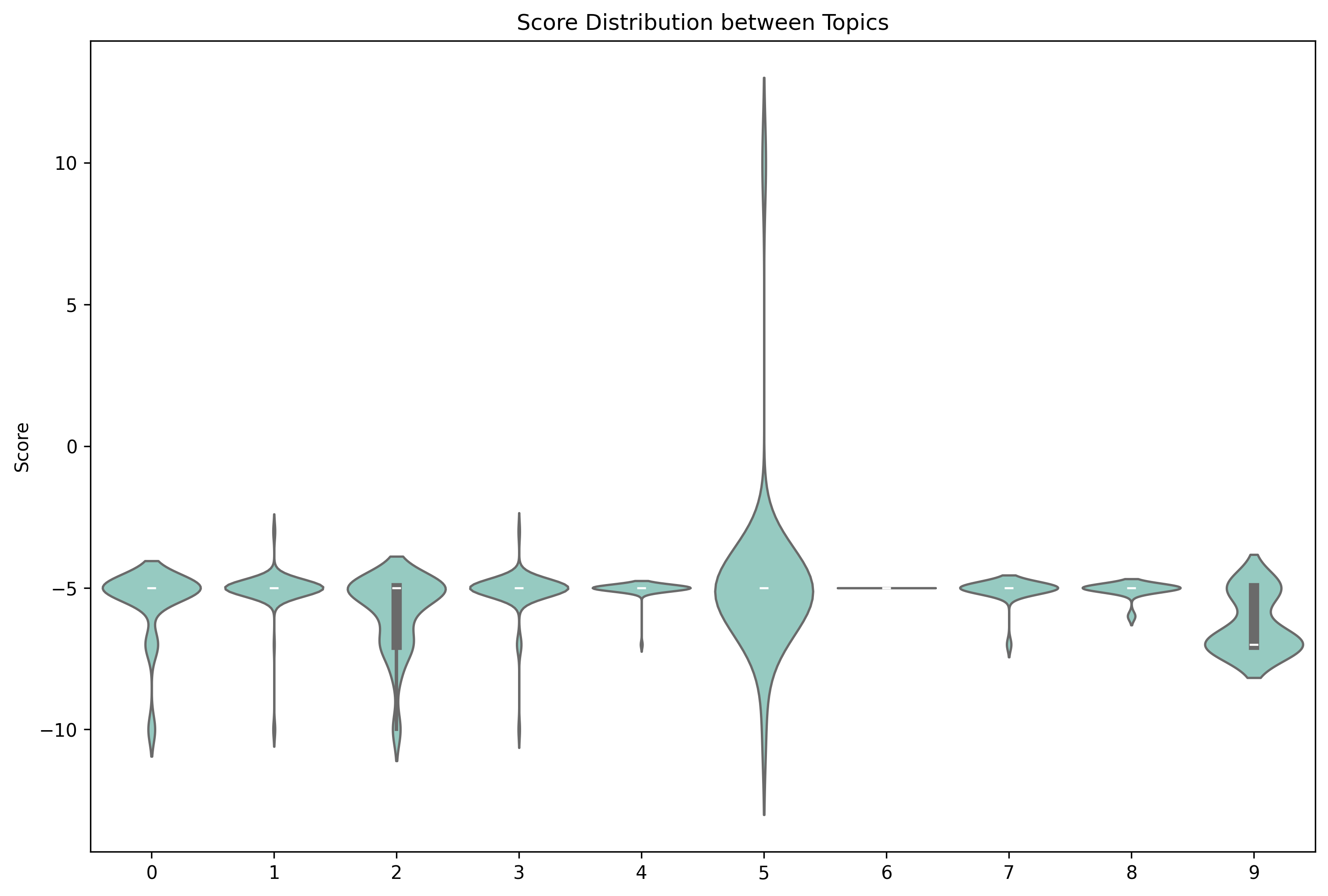}
 \caption{Distribution of cluster scores of GLM-4.}
 \label{img:glm4_reasoning}
\end{minipage}
\end{figure}



\begin{figure}[htbp]
\begin{minipage}{0.48\linewidth}
 \centering
 \includegraphics[width=\linewidth]{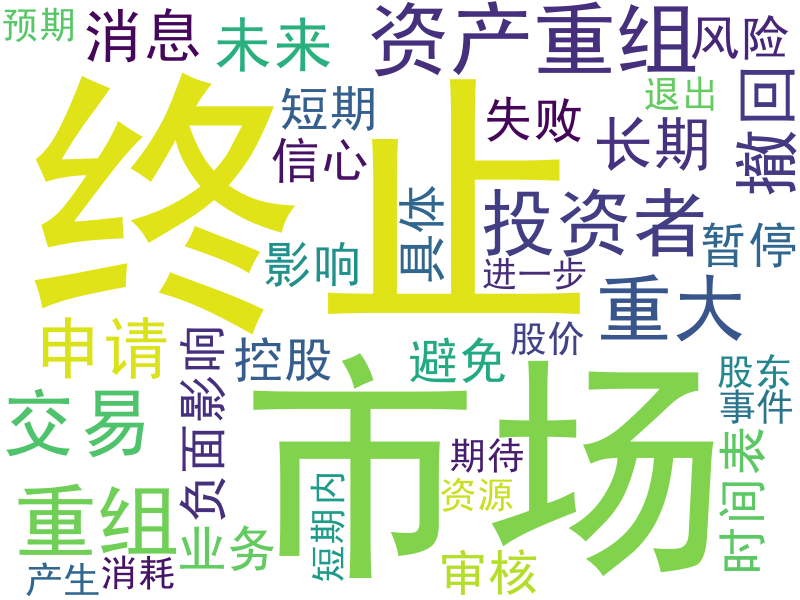}
 \caption{The wordcloud of GLM-4}
 \label{img:glm_wordcloud}
\end{minipage}
\hfill 
\begin{minipage}{0.48\linewidth}
 \centering
 \includegraphics[width=\linewidth]{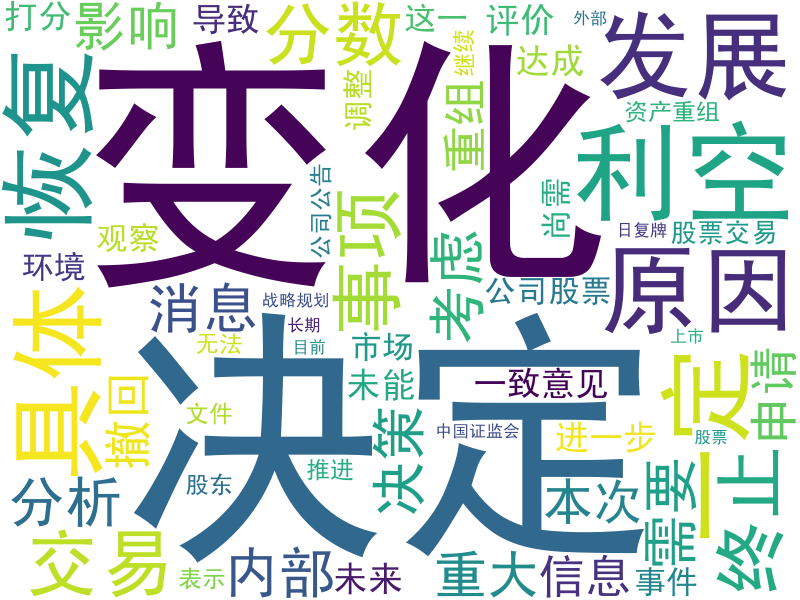}
 \caption{The wordcloud of Baichuan2-7B.}
 \label{img:baichuan_wordcloud}
\end{minipage}
\end{figure}

We further analyze the reasoning texts of the best performing GLM-4 and the worst performing Baichuan2-7B, clustering them into 10 groups with 10 keywords in each group. The key vocabulary of the two models will be summarized and a word cloud will be drawn. The results are shown in \autoref{img:glm_wordcloud} and \autoref{img:baichuan_wordcloud}.

\subsection{Analysis of Interactions}
\label{sec:a73}
We process and analyze the information related to the interaction between the company and shareholders in a similar way.

\foreach \i in {0,2,4,...,8} { 
    \begin{figure}[]
    \begin{minipage}{0.48\linewidth}
        \centering
        \includegraphics[width=\linewidth]{fig/interaction/interaction_\i.png}
        \caption{Distribution of the score in interaction\the\numexpr\i+1\relax.}
        \label{img:interaction_distribution\i}
    \end{minipage}
    \hfill 
    \begin{minipage}{0.48\linewidth}
        \centering
        \includegraphics[width=\linewidth]{fig/interaction/interaction_\the\numexpr\i+1\relax.png}
        \caption{Distribution of the score in interaction\the\numexpr\i+2\relax.}
        \label{img:interaction_distribution\the\numexpr\i+1\relax}
    \end{minipage}
    \end{figure}
}

\clearpage
\subsection{Analysis of Risk-preference Questions}
\label{sec:a74}


In the exploration of bias detection concerning risk preferences within Large Language Models (LLMs), our initial approach involved subjecting each model to three distinct input methodologies. The outcomes of these initial tests, aimed at gauging the models' inherent risk preferences, are meticulously documented in \autoref{tab:model_risk_preferences}. This foundational analysis sets the stage for more nuanced investigations into model behaviors under specific conditions.

Subsequently, our focus shifted to the models' adherence to explicit instructions regarding risk aversion. By inputting the directive "You are a risk averse person," we were able to quantify each model's compliance through the risk averse ratio, the details of which are encapsulated in \autoref{tab:instruct_risk}. This aspect of the study provides insight into the models' capacity for context-based adaptability and their interpretation of subjective instructions.

Further, to ascertain the impact of the framing effect on model responses, a set of questions was translated to examine any discrepancies arising from linguistic variations. The findings from this segment of the study, highlighting the influence of translation on model outputs, are presented in \autoref{tab:translation_difference}. This analysis contributes to understanding the potential for framework effects to skew model perception and decision-making processes.

Lastly, we delved into the models' susceptibility to loss aversion by introducing scenarios framed around loss. The extent of loss aversion bias manifesting in the models' responses was rigorously analyzed, with the summarized results being showcased in \autoref{tab:Loss_Aversion_Bias}. Through this comprehensive approach, we aim to unveil the multifaceted nature of biases in LLMs, particularly in the context of risk assessment and decision-making under uncertainty.

\begin{table}[htbp]
\centering
\caption{Model Instruct Risk-aversion Performance Comparison}
\begin{tabular}{l l}
\hline
\textbf{Model} & \textbf{Risk-aversion (\%)} \\
\hline
GPT-4 & 89.5 \\
Qwen-max & 83.5 \\
GLM-4 & 88.0 \\
Qwen-72B & 79.0 \\
ChatGLM3-Turbo & 62.5 \\
Xuanyuan-70B & 59.5 \\
Qwen-14B & 66.0 \\
InternLM2-7B & 42.5 \\
Baichuan2-13B & 44.0 \\
FinQwen & 45.0 \\
Xuanyuan-13B & 43.0 \\
ChatGLM3-6B & 53.0 \\
InternLM2-20B & 53.0 \\
Qwen-7B & 52.0 \\
Baichuan2-7B & 38.0 \\
GPT-3.5 & 37.5 \\
ChatGLM2-6B & 35.0 \\
\hline
\label{tab:instruct_risk}
\end{tabular}
\end{table}

\begin{table}[htbp]
\centering
\caption{Models Translation Prompt Differences Comparison}
\begin{tabular}{l l}
\hline
\textbf{Model} & \textbf{Difference (\%)} \\
\hline
GPT-4 & 23.5 \\
ChatGLM3-6B & 25.0 \\
Qwen-max & 28.5 \\
GLM-4 & 28.0 \\
Xuanyuan-70B & 36.0 \\
GPT-3\_5 & 33.0 \\
Qwen-7B & 42.0 \\
InternLM2-7B & 43.5 \\
ChatGLM3-Turbo & 46.5 \\
FinQwen & 49.0 \\
ChatGLM2-6B & 51.0 \\
Xuanyuan-13B & 51.5 \\
Baichuan2-13B & 54.5 \\
InternLM2-20B & 48.5 \\
Qwen-72B & 56.0 \\
Baichuan2-7B & 59.0 \\
Qwen-14B & 65.0 \\
\hline
\label{tab:translation_difference}
\end{tabular}
\end{table}

\begin{table}[htbp]
\centering
\caption{Model Loss Aversion Bias Comparison}
\begin{tabular}{l l}
\hline
\textbf{Model} & \textbf{Risk-aversion (\%)} \\
\hline
Qwen-14B & 51.0 \\
GPT-3\_5 & 52.0 \\
FinQwen & 57.5 \\
Baichuan2-7B & 58.5 \\
InternLM2-7B & 60.5 \\
Qwen-max & 62.0 \\
Baichuan2-13B & 63.0 \\
InternLM2-20B & 63.0 \\
ChatGLM2-6B & 61.5 \\
Qwen-72B & 65.5 \\
Xuanyuan-13B & 66.5 \\
GLM-4 & 69.0 \\
ChatGLM3-Turbo & 74.0 \\
Xuanyuan-70B & 74.5 \\
Qwen-7B & 72.5 \\
ChatGLM3-6B & 75.0 \\
GPT-4 & 84.0 \\

\hline
\label{tab:Loss_Aversion_Bias}
\end{tabular}
\end{table}

\begin{table}[htbp]
\centering
\caption{LLMs risk preference statistics}
\label{tab:model_risk_preferences}
\begin{tabular}{lcccc}
\toprule
Model & Method & Risk Averter & Risk Neutral& Risk Lover \\
\midrule
Baichuan2-7B & Direct & 67 & 35 & 98 \\
Baichuan2-7B & Instruct & 76 & 24 & 100 \\
Baichuan2-7B & Translation & 67 & 58 & 75 \\
Qwen-72B & Direct & 81 & 79 & 40 \\
Qwen-72B & Instruct & 160 & 32 & 8 \\
Qwen-72B & Translation & 85 & 45 & 70 \\
Qwen-14B & Direct & 52 & 46 & 102 \\
Qwen-14B & Instruct & 134 & 27 & 39 \\
Qwen-14B & Translation & 112 & 15 & 73 \\
GLM-4 & Direct & 88 & 32 & 80 \\
GLM-4 & Instruct & 178 & 12 & 10 \\
GLM-4 & Translation & 92 & 39 & 69 \\
ChatGLM2-6B & Direct & 73 & 13 & 114 \\
ChatGLM2-6B & Instruct & 70 & 17 & 113 \\
ChatGLM2-6B & Translation & 101 & 23 & 76 \\
ChatGLM3-6B & Direct & 100 & 18 & 82 \\
ChatGLM3-6B & Instruct & 107 & 21 & 72 \\
ChatGLM3-6B & Translation & 99 & 28 & 73 \\
Xuanyuan-70B & Direct & 99 & 24 & 77 \\
Xuanyuan-70B & Instruct & 120 & 24 & 56 \\
Xuanyuan-70B & Translation & 90 & 20 & 90 \\
InternLM2-7B & Direct & 71 & 70 & 59 \\
InternLM2-7B & Instruct & 86 & 69 & 45 \\
InternLM2-7B & Translation & 81 & 63 & 56 \\
Baichuan2-13B & Direct & 76 & 16 & 108 \\
Baichuan2-13B & Instruct & 89 & 32 & 79 \\
Baichuan2-13B & Translation & 61 & 39 & 100 \\
Qwen-7B & Direct & 95 & 27 & 78 \\
Qwen-7B & Instruct & 105 & 28 & 67 \\
Qwen-7B & Translation & 106 & 24 & 70 \\
InternLM2-20B & Direct & 76 & 76 & 48 \\
InternLM2-20B & Instruct & 108 & 63 & 29 \\
InternLM2-20B & Translation & 73 & 78 & 49 \\
ChatGLM3-Turbo & Direct & 98 & 45 & 57 \\
ChatGLM3-Turbo & Instruct & 127 & 48 & 25 \\
ChatGLM3-Turbo & Translation & 62 & 85 & 53 \\
GPT-3.5 & Direct & 54 & 35 & 111 \\
GPT-3.5 & Instruct & 75 & 24 & 101 \\
GPT-3.5 & Translation & 56 & 27 & 117 \\
FinQwen & Direct & 65 & 42 & 93 \\
FinQwen & Instruct & 91 & 34 & 75 \\
FinQwen & Translation & 101 & 46 & 53 \\
GPT-4 & Direct & 118 & 41 & 41 \\
GPT-4 & Instruct & 181 & 11 & 8 \\
GPT-4 & Translation & 128 & 39 & 33 \\
Xuanyuan-13B & Direct & 83 & 26 & 91 \\
Xuanyuan-13B & Instruct & 86 & 24 & 90 \\
Xuanyuan-13B & Translation & 106 & 13 & 81 \\
Qwen-max & Direct & 74 & 89 & 37 \\
Qwen-max & Instruct & 169 & 26 & 5 \\
Qwen-max & Translation & 99 & 62 & 39 \\
\bottomrule
\end{tabular}
\end{table}

\end{document}